\documentclass[lettersize,journal]{IEEEtran}
\usepackage{amsmath,amsfonts}
\usepackage{algorithmic}
\usepackage{algorithm}
\usepackage{array}
\usepackage{textcomp}
\usepackage{stfloats}
\usepackage{url}
\usepackage[hidelinks]{hyperref}  
\usepackage{verbatim}
\usepackage{graphicx}
\usepackage{cite}
\usepackage{caption}
\usepackage{xcolor}         
\usepackage{amsmath}
\usepackage{tikz} 
\usepackage{cite}
\usepackage{subcaption}

\usepackage[utf8]{inputenc} 
\usepackage[T1]{fontenc}    
\usepackage{url}            
\usepackage{booktabs}       
\usepackage{amsfonts}       
\usepackage{nicefrac}       
\usepackage{microtype}      
\usepackage{xcolor}         
\usepackage[export]{adjustbox}
\usepackage{threeparttable}
\usepackage{multirow}
\usepackage{enumitem}
\usepackage{newfloat}
\usepackage{listings}
\usepackage{colortbl}
\usepackage{amsfonts}
\usepackage{caption}
\usepackage{wrapfig}
\usepackage{amssymb}
\usepackage{pifont}
\usepackage{float}
\usepackage{bm}
\usepackage{placeins}

\usepackage{tcolorbox}
\tcbuselibrary{theorems}
\definecolor{lightgray}{gray}{.9}
\definecolor{deepgray}{gray}{.8}
\tcbset{highlight math/.append style={left=0mm,right=0mm,top=0mm,bottom=0mm, colframe=white}}

\definecolor{Periwinkle}{RGB}{204, 204, 255} 

\definecolor{mygray}{gray}{.9}
\definecolor{mygreen}{RGB}{93,173,85}
\definecolor{mywarning}{RGB}{233,144,61}

\definecolor{DarkBlue}{RGB}{64,101,149}
\definecolor{azure}{rgb}{0.0, 0.5, 1.0}
\definecolor{gray}{rgb}{0.3, 0.3, 0.3}
\definecolor{DarkGreen}{RGB}{42,110,63}

\makeatletter
\newcommand{\thickhline}{%
    \noalign {\ifnum 0=`}\fi \hrule height 1pt
    \futurelet \reserved@a \@xhline
}



\newcolumntype{x}[1]{>{\centering\arraybackslash}p{#1pt}}
\newcolumntype{I}{!{\vrule width 1pt}}

\usepackage[capitalize]{cleveref}
\crefname{proposition}{Prop.}{Props.}
\crefname{section}{Sec.}{Secs.}
\crefname{table}{Tab.}{Tabs.}

\usepackage{xspace}
\makeatletter
\DeclareRobustCommand\onedot{\futurelet\@let@token\@onedot}
\def\@onedot{\ifx\@let@token.\else.\null\fi\xspace}

\title{Adversarial Curriculum Graph-Free Knowledge Distillation for Graph Neural Networks}

\author{
\begin{tabular}{c}
\textbf{Yuang~Jia}$^{1}$ \quad
\textbf{Xiaojun~Shan}$^{2}$ \quad
\textbf{Jun~Xia}$^{3*}$\thanks{%
    \rule[0.5ex]{8cm}{0.5pt}\\[0.5ex]
    \hspace*{1.5em}*Corresponding author
} \quad
\textbf{Guancheng~Wan}$^{4}$ \quad
\textbf{Yuchen~Zhang}$^{1}$ \\
\textbf{Wenke~Huang}$^{4}$ \quad
\textbf{Mang~Ye}$^{4}$ \quad
\textbf{Stan~Z.~Li}$^{3}$ \\
$^{1}$University of Electronic Science and Technology of China \quad
$^{2}$University of California San Diego \\
$^{3}$Westlake University \quad
$^{4}$Wuhan University \\
\end{tabular}
}

\IEEEaftertitletext{\vspace{-3em}} 

\begin{document}
\maketitle
\begin{abstract}
Data-free Knowledge Distillation (DFKD) is a method that constructs pseudo-samples using a generator without real data, and transfers knowledge from a teacher model to a student by enforcing the student to overcome dimensional differences and learn to mimic the teacher's outputs on these pseudo-samples. In recent years, various studies in the vision domain have made notable advancements in this area. However, the varying topological structures and non-grid nature of graph data render the methods from the vision domain ineffective. Building upon prior research into differentiable methods for graph neural networks, we propose a fast and high-quality data-free knowledge distillation approach in this paper. Without compromising distillation quality, the proposed graph-free KD method (ACGKD) significantly reduces the spatial complexity of pseudo-graphs by leveraging the Binary Concrete distribution to model the graph structure and introducing a spatial complexity tuning parameter. This approach enables efficient gradient computation for the graph structure, thereby accelerating the overall distillation process. Additionally, ACGKD eliminates the dimensional ambiguity between the student and teacher models by increasing the student's dimensions and reusing the teacher's classifier. Moreover, it equips graph knowledge distillation with a CL-based strategy to ensure the student learns graph structures progressively. Extensive experiments demonstrate that ACGKD achieves state-of-the-art performance in distilling knowledge from GNNs without training data. 
\end{abstract}


\vspace{5pt}
\section{Introduction}
Data-Free Knowledge Distillation (DFKD) \cite{binici2022robust, do2022momentum, fang2021contrastive, patel2023learning, tran2024text, yin2020dreaming, Li2022HowTT,yu2023data,2022fine, chen2023best} is an effective and emerging  approach that generates pseudo-samples for knowledge transfer from a teacher network to a student in the absence of real data. Its core principle involves feeding the samples generated by the generator to the pretrained teacher model, iterating to maximize the class-conditional probability, and continuously updating the generator's parameters to make the pseudo-samples gradually realistic. Due to the inaccessibility of most privacy-related information, such as social and medical data in various areas of life, DFKD has become an increasingly popular approach in various domains.

\vspace{1.5pt}

DFKD methods \cite{r21,cudfkd,ctkd,micaelli2019zero,tran2024nayer,liao2024impartial} generally use a generator to synthesize images. In this setup, the student strives to match the teacher’s predictions on the pseudo images, while the generator aims to create samples that maximize the discrepancy between the student’s and teacher’s predictions. This adversarial training approach has achieved remarkable results in vision distillation tasks. In recent years, driven by the need to synthesize high-quality graph data, such as biological protein structures and social networks, some studies have shifted their focus to the field of graph-free knowledge distillation, which addresse the unique challenges posed by graph data structures. For example, Deng et al.~\cite{gfkd} applies the Bernoulli distribution to model the graph's topological structure, allowing gradient updates for discrete graph structures. Building on this, Zhuang et al.~\cite{gfad} introduces adversarial training into the field, improving the distillation effect by training a separate generator at the cost of increased time for graph generation. Zhu et al.~\cite{fedtad}, on the other hand, generates pseudo-graphs using class-specific information uploaded by clients and performs knowledge distillation for student models on the server side.
\begin{figure}[t]
    \centering
    \includegraphics[width=\columnwidth]{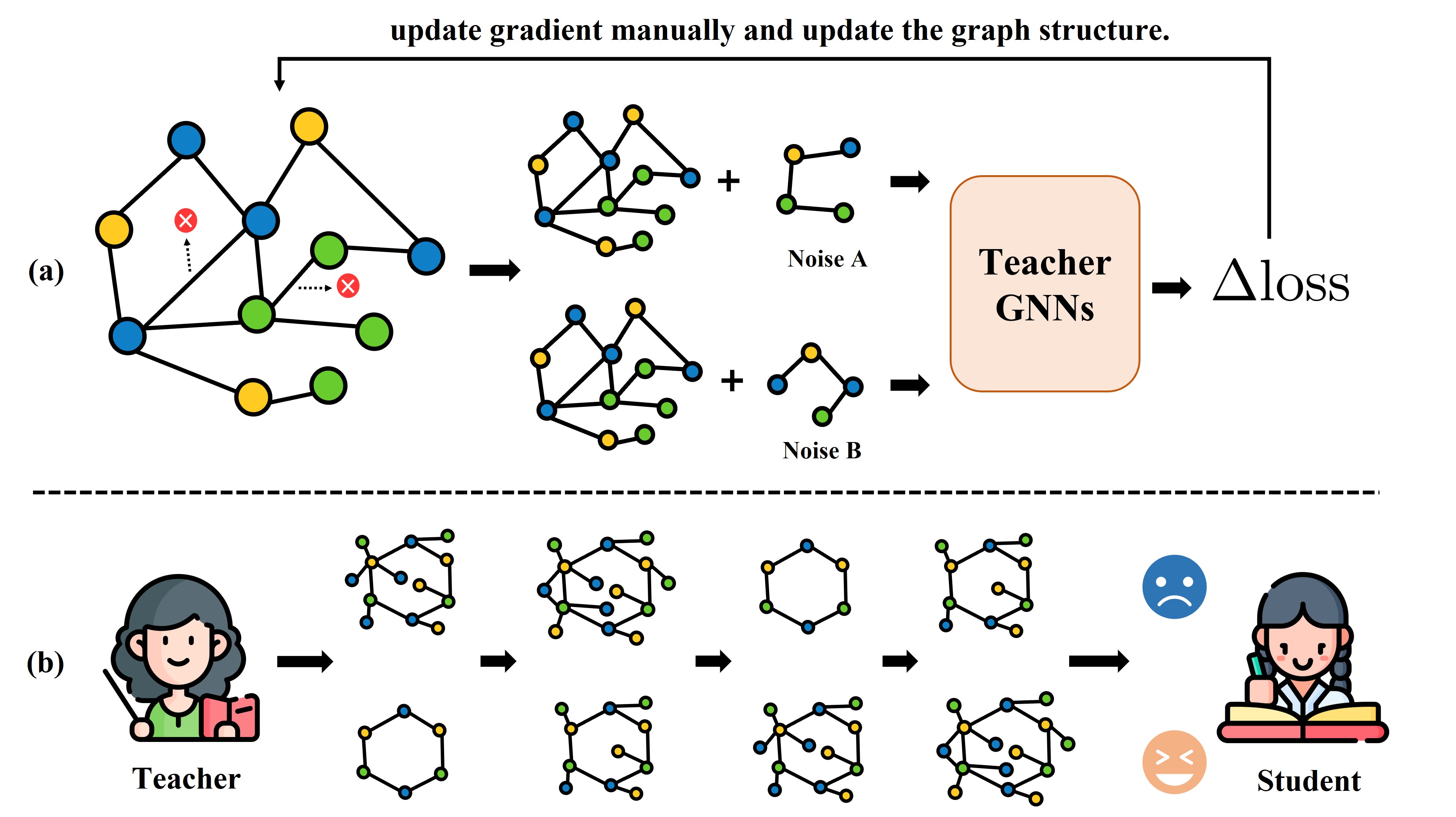}  
    \captionsetup{justification=raggedright, singlelinecheck=false}  
    \caption{\normalsize Problem illustration. (a) During the gradient compu \\ -tation of a graph structure, the Bernoulli-based parameter optimization method introduces two noise terms. Both are input into the teacher model, and gradients are manually computed from the difference between the two losses. (b) An inappropriate order of knowledge transfer can confuse the student while resulting in low sample utilization.}  
    \vspace{-15pt}
    \label{problem}  
\end{figure}

Despite their remarkable effectiveness, considering that distillation should prioritize both efficiency and effectiveness, existing graph-free knowledge distillation methods do not restrict the complexity of graph structures and commonly model them using the Bernoulli distribution. As shown in Fig. \ref{problem}, this non-differentiable approach requires additional structural gradient computation, significantly increasing the time required to generate these samples, and the gradient estimation method \cite{sutton1999policy} it employs introduces high variance, leading to unstable training. Moreover, the binary nature of the Bernoulli distribution (i.e., the presence or absence of edges) is highly unfavorable for Graph Neural Networks, as the lack of soft connections raises the risk of losing critical edges. This raises the first issue: \hypertarget{Q1}{\textbf{\uppercase\expandafter{\romannumeral1})}} \textit{Is it possible to modify the graph modeling approach to reduce spatial complexity and keep critical edges while preserving the quality of pseudo-data?} Furthermore, the spatial complexity of these generated pseudo-graphs varies, and the generation order is entirely random, which contradicts the natural principle in real life where a teacher typically imparts knowledge to students in a progression from easy to hard. Just as illustrated in Fig. \ref{problem}, this undoubtedly leads to inefficiency of pseudo-graphs, as the student model cannot comprehend complex knowledge in the early stages. Additionally, this also affects the student model's final knowledge acquisition, specifically the correctness of the learned parameters. Therefore, we consider the second issue: \hypertarget{Q2}{\textbf{\uppercase\expandafter{\romannumeral2})}} \textit{What is the best way to utilize the generated pseudo-graphs to maximize their effectiveness?} These two issues motivate our study in this paper.

To address the challenges mentioned above, we propose ACGKD, a fast and efficient graph-free knowledge distillation method. In general, ACGKD proposes a strategy for efficiently computing graph structural gradients and significantly reducing the spatial complexity of generated graphs, while also establishing a comprehensive curriculum learning system for the student model. Specifically, \ding{182} to efficiently compute structural gradients and retain potential key edges, we model the pseudo-graphs using the Binary Concrete distribution \cite{2016concrete}, which preserves soft edge information and converts discrete graph structures into continuous ones, thus enabling direct gradient backpropagation. Inspired by research on scale-free networks \cite{tkde-pandey2017parametric, tkde-xia2022cengcn}, we also introduce a trainable spatial complexity parameter \(\xi\), which significantly reduces the graph's complexity based on feedback from the teacher model. \ding{183} We introduce curriculum learning into the graph KD field, gradually increasing the difficulty of the pseudo-graphs and dynamically controlling the student model’s focus to enable effective learning from simple to complex tasks. We also apply a dynamic temperature parameter, which is trained adversarially to maximize the distillation loss, thereby gradually increasing the difficulty for the student. It serves as an additional components to curriculum learning.

We conduct detailed experiments on six graph datasets with different student-teacher model architectures demonstrating that ACGKD significantly reduces spatial complexity and improves overall distillation performance. To summarize, we make the following key contributions in this paper:
\begin{itemize}
\item We propose an efficient graph-free distillation approach, ACGKD, which models graph topology using the Binary Concrete Distribution, thus preserving potential key edge information and accelerating structural gradient computation. Additionally, we introduce a spatial complexity parameter during the generation process to simplify the graph structure without affecting distillation quality, ultimately reducing the overall time for knowledge distillation.
\item We incorporate curriculum learning and dynamic temperature adjustments into graph knowledge distillation, enabling the student model to perform adversarial and effective learning, progressing from simple to complex samples, while also making full use of samples.
\item Extensive experiments show that ACGKD outperforms existing graph-free distillation methods, and various student-teacher combinations validate ACGKD's generalizability.
\end{itemize}

\section{RELATED WORK}
\subsection{Data-Free Knowledge Distillation}
\label{dfkd}
DFKD methods typically synthesize images to transfer knowledge from a teacher to a student model in the field of computer vision. For instance, Yin et al.~\cite{yin2020dreaming} uses the teacher's batch normalization to optimize random noise, while Fang et al.\cite{fang2021contrastive}, Yu et al.~\cite{yu2023data} refine noise iteratively, and Do et al.\cite{r2}, Patel et al.~\cite{do2022momentum}, fang et al.~\cite{patel2023learning} employ a generator to capture data distribution. The FM method \cite{r7} accelerates this process by integrating a meta-generator. The generated data are used to jointly train the generator and student in an adversarial setup \cite{r21,cudfkd}, with the student mimicking the teacher’s predictions while the generator amplifies their differences. Recently, the application of graph data has grown rapidly. However, in many practical scenarios, obtaining high-quality real-world graph data is often costly or impractical. This makes it necessary to generate graph data through data-free methods. Traditional DFKD approaches, primarily designed for Euclidean data in the visual domain, fail to handle graph data effectively due to its structural complexity and irregularity. Thus, there is a strong need to develop data-free KD approaches specifically tailored for GNNs. A notable example is \cite{gfkd}, which facilitates knowledge transfer by representing graph topological structures using a multivariate Bernoulli distribution. Building on this, Zhuang et al.~\cite{gfad} applies the generator-student adversarial approach in the graph domain and has made some progress. Moreover, Zhu et al.~\cite{fedtad} combines federated learning to generate pseudo-graphs using class-specific information uploaded by clients, and performs knowledge distillation for the student models on the server side. However, the computational cost of generating pseudo-graphs in these methods is high, and the difficulty of the generated pseudo-graphs varies irregularly.

\subsection{Curriculum Learning}
Bengio et al. \cite{r2} introduce Curriculum Learning (CL), a strategy to improve model performance by progressively incorporating samples from easy to hard. Self-paced learning \cite{r25} extends this by evaluating sample difficulty based on training loss, allowing the model to adjust its curriculum dynamically. Later studies like \cite{r18,r17,r55} establish metrics like sample diversity \cite{r17} and prediction consistency \cite{r55} to guide curriculum design. Empirical studies, such as MentorNet \cite{r19} and Co-teaching \cite{r15}, demonstrate that CL enhances generalization under noisy conditions, while theoretical work \cite{r11} highlight its denoising effect by reducing the focus on noisy samples. Research \cite{r2,r35,r46,r13,r24} recognize CL’s role in expediting non-convex optimization and accelerating convergence in early training phases. Additionally, Zhao et al.~\cite{tkde-zhao2024ccml} and Wang et al.~\cite{tkde-wang2024unified} use curriculum learning to enhance the meta-learning framework, effectively addressing data bias issues in information retrieval systems. Despite significant achievements, most existing CL strategies are designed for independent data types, like images, with limited research on adapting CL strategies for samples with dependencies. Some attempts on graph-structured data \cite{r26,r21,r29}, such as \cite{r44,r5,r45,r29}, merely treat nodes as independent samples and apply CL methods designed for independent data, neglecting the holistic information embedded within the graph structure. Additionally, these methods primarily rely on heuristic sample selection strategies \cite{r5,r45,patel2023learning}, which significantly restrict the generalizability of these approaches. Unlike the methods mentioned above, our approach employs an adaptive generation strategy that gradually increases the complexity of generated pseudo-graphs for both nodes and structure, enabling the student model to assimilate knowledge more effectively. Additionally, we introduce a temperature MLP layer to encourage adversarial learning, further enhancing the student model's learning capabilities.

\vspace{-1pt}
\section{Method}
\subsection{Overview}
\begin{figure*}[t]
    \centering
    \includegraphics[width=\textwidth]{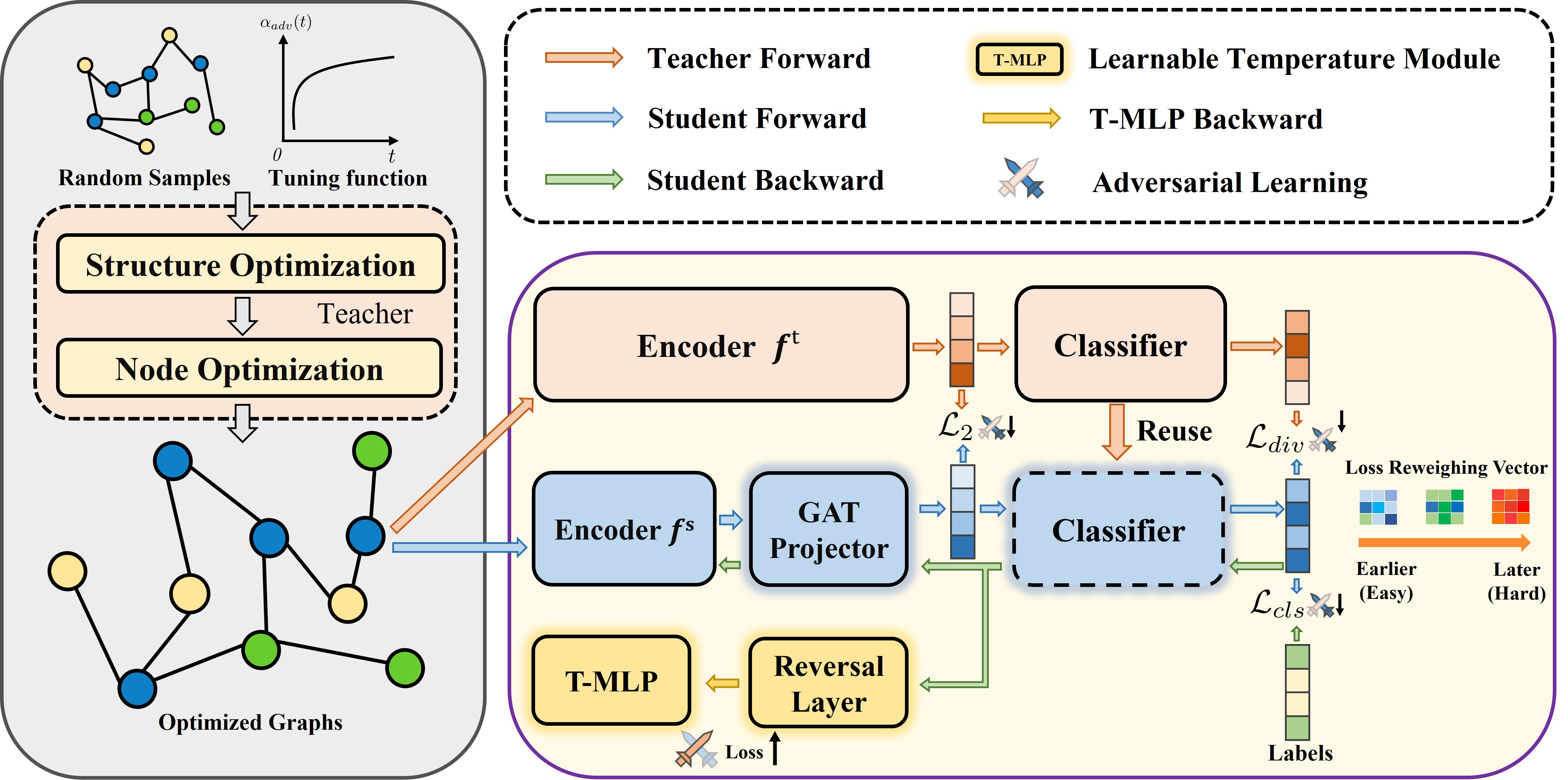}  
    \captionsetup{justification=centering}  
    \caption{\normalsize The overall architecture of ACGKD.}  
    \vspace{-10pt}
    \label{pipeline}  
\end{figure*}

Fig. \ref{pipeline} shows the overview of our proposed architecture. We feed randomly initialized node feature and graph structure parameters into the teacher model. These parameters are used to establish the Binary Concrete distribution. At different stages of data generation, ACGKD controls the pseudo-graphs generated by the teacher model from easy to hard. The student model receiving the pseudo-graphs produces an output with dimension different from the teacher model, so we use a projector, which is at a relatively small cost yet ensures accurate dimension alignment. Then we reuse the classifier of the teacher model on the student model for comparison with ground truth labels. The dynamic vectors in this process are used to control the student model's learning focus so that key parameters are learned early and the more difficult parameters are learned later. Furthermore, we add a learnable temperature module to encourage adversarial learning. The details of each part will be introduced in the following sections.

\vspace{-8pt}
\subsection{Pseudo-graph Generation with Reduced Spatial Complexity}
Assume that the teacher GNN \(T(\cdot)\) is trained on the dataset \((X, Y)\) by minimizing the cross-entropy loss function \(\mathcal{L}_{CE}\):

\begin{equation}
    \mathcal{L}_{C E}=\mathcal{C}\left(Y, T_{W}(H, A)\right) \tag{1}
    \label{eq1}
\end{equation}
where \(X\) represents the graph data, \(Y\) represents the labels, \(W\) represents the parameters of \(T\) and \(\mathcal{C}\)  denotes the loss function, such as cross-entropy or mean square error.

From the Bayesian perspective \cite{bernardo2009bayesian}, learning \(W\) can be viewed as maximizing the class-conditional probability \(p(Y\mid N, S, W)\), i.e., \(\operatorname{argmax}_{W} \log p(Y \mid N, S, W)\), in which \(N\) denotes the node features of the graphs \(X\); \(S\) represents the graph structure information of \(X\), which can be expressed as adjacency matrices composed of 0s and 1s. When the training data is unavailable but the teacher model \(W\) is known, pseudo-samples can be generated by optimizing the input (i.e., \(N\) and \(S\)) for knowledge transfer. However, this approach is not applicable to GNNs because \(\mathcal{L}_{CE}\) is non-differentiable with respect to the graph structure \(S\).

To transform the topological graph into a structure that can be quickly differentiated, inspired by \cite{2016concrete}, we model the graph's topology using the Binary Concrete distribution (the initial graphs are randomly generated in our method). This not only makes the graph structure differentiable but also avoids the heavy computations in \cite{gfkd}. Specifically, this distribution converts binary edges to continuous values in (0,1) range:

\begin{equation}
    s_{i j}=\sigma\left(\frac{\log \alpha_{i j} + G_{i j}}{\lambda}\right)
    \tag{2}
    \label{eq2}
\end{equation}
where \(\log \alpha_{i j}\) is the logarithmic value of the original binary probability \(\alpha_{i j}\) representing the weight of the edge, which can be obtained during initialization. \(G_{i j} \sim\) Gumbel (0,1) is Gumbel noise that facilitates the discrete-to-continuous transition. \(\lambda\) is a temperature-like parameter that controls the degree of relaxation in the sigmoid operation.

For a batch of graphs, their structures \(S\) are independently sampled from the joint distribution \(\prod_{i=1}^{m} P_{\alpha_{i}}\left(S_{i}\right)\), where \(m\) is the number of graphs and \(P_{\alpha_{i}}\left(S_{i}\right)\) represents the Binary Concrete distribution for the structure of graph \(i\) parameterized by \(\alpha_{i}\), which controls the soft probabilities for the existence of edges. Instead of minimizing the non-differentiable loss function (\ref{eq1}), we synthesis pseudo-graphs by minimizing the following expectation:

\begin{equation}
    \mathcal{L}_{N, S}=\underset{S \sim P(S)}{\mathbb{E}}\left[\mathcal{C}\left(Y, T_{W}(N, S)\right)+ \rho * \mathcal{R}\right]\tag{3}
    \label{eq3}
\end{equation}
where \(P(S)=\prod_{i=1}^{m} P_{\alpha_{i}}\left(S_{i}\right)\), \(N\) denotes the node feature parameters, \(Y\) is a set of randomly sampled labels, and \(R\) represents the regularizers for various priors on the target task data(e.g., normalization). \(\rho\) is a balancing weight. Minimizing (\ref{eq3}) with respect to \(N\) and \(S\) generates pseudo-graphs that maximize the class probability of the GNN's output. We omit \(W\) in \(T_W\) in the following as it is already known for the pre-trained teacher GNN.

To further reduce the spatial complexity of the generated pseudo-graphs and accelerate the distillation process, we introduce a trainable spatial complexity parameter \(\xi\) during data generation. This parameter represents the number of nodes reduced from the original graph. This parameter reduces the size of the adjacency matrix, significantly lowering the spatial complexity and the total number of feature parameters. Assuming the original number of nodes is \(n\), for undirected graphs, the number of node parameters is reduced from \(n^2\) to \((n - \xi)^2\); for directed graphs, it is reduced from \(\frac{n(n+ 1)}{2}\) to \(\frac{(n - \xi)(n - \xi + 1)}{2}\). As spatial complexity decreases quadratically, both pseudo-graphs generation time and knowledge distillation time are significantly reduced. After introducing parameter \(\xi\), equation (\ref{eq3}) becomes:

\begin{equation}
    \mathcal{L}_{N, S, \xi}=\underset{S \sim P(S)}{\mathbb{E}}\left[\mathcal{C}\left(Y, T(N, S, \xi)\right)+ \rho * \mathcal{R}\right]\tag{4}
    \label{eq4}
\end{equation}

To minimize the objective in (\ref{eq4}), it is necessary to compute the gradients with respect to \(N\), \(S\), and \(\xi\). Given that we model the graph using the Binary Concrete distribution, computing the gradients for these three parameters becomes straightforward by sampling from \(P(S)\):

\begin{equation}
    \begin{aligned}\nabla \mathcal{L}_{N, S, \xi} & =\nabla\underset{S \sim P(S)}{\mathbb{E}}[\mathcal{C}(Y, T(N, S, \xi)) + \rho * \mathcal{R}] \\& =\underset{S \sim P(S)}{\mathbb{E}} \nabla[\mathcal{C}(Y, T(N, S, \xi))+ \rho * \mathcal{R}] \\& =\frac{1}{Q} \sum_{i=1}^{Q} \nabla\left[\mathcal{C}\left(Y, T\left(N, S^{i}, \xi\right)\right)+ \rho * \mathcal{R}\right]\end{aligned}\tag{5}
    \label{eq5}
\end{equation}
where \(S^{i} \sim P(S)\) are i.i.d. samples, \(Q\) is the number of samples. After multiple rounds of backpropagation and gradient updates, the randomly initialized graphs evolve into ones with clearer node features and structure.

\vspace{-8pt}
\subsection{Dimension Alignment and Reuse of Teacher's Classifier }

It is generally believed that in tasks that deal with different data distributions, a common approach is to use similar feature extraction layers in shallow layers to extract some general information, and in deeper layers to extract task-specific information. In current data knowledge distillation tasks, student models need to separately train their own classifiers. This results in the student model needing to additionally learn deeper network parameters to extract task-specific information. However, since the number of layers of the teacher model is often deeper, its classifier can not only be used for final classification but also contains implicit knowledge of the graph's topological structure. Therefore, in order to allow the student model to utilize this implicit knowledge concisely and efficiently, and to avoid retraining unnecessary classifiers that may not be effective, we reuse the classifier of the teacher model. Before reusing the classifier, the student model’s intermediate output dimensions should be aligned with the teacher model through a projection operation to eliminate dimensional ambiguity.

We assume that the intermediate output of the student model before projection is recorded as \(h^{s}\). The input feature of the i-th node is denoted as \(h^{s}_i\). In our approach, we ultimately chose the Graph Attention Network (GAT) \cite{velivckovic2017graph} as the projector, since it captures varying node relationships effectively through adaptive attention weights, thus enhancing the learning of more complex graph structures:
\begin{equation}
h^{'}_i=\frac{1}{K} \sum_{k=1}^{K} \sigma\left(\sum_{j \in \mathcal{N}_{i}} \alpha_{i j}^{k} {W}^{k} h_{j}^{s}\right) \tag{6}
\label{eq6}
\end{equation}
where \(\sigma\) denotes the nonlinear activation function. \(h_{j}^{s}\) is the original feature representation of node \(j\). \(\alpha_{i j}^{k}\) is the normalized attention coefficient between node \(i\) and node \(j\) under the k-th attention head.
\(W^{k}\) is the weight matrix of the k-th attention head and \(h^{'}_i\) denotes the new feature representation of node \(i\) after GAT projection.

For the teacher model’s classifier, we reuse a simple linear and dropout layer to convert the hidden dimensions into the corresponding output dimensions required for the task.

\vspace{-6pt}
\subsection{Curriculum Learning of Graph Data}
We imitate and extend two plug-and-play modules on curriculum learning in \cite{cudfkd}. These two modules are used respectively when the teacher model generates pseudo-graphs and the student model is trained. In our method, the parameters of the teacher model are frozen when generating pseudo-graphs, and our goal is to generate easy to hard pseudo-graphs by optimizing the graph structure. So unlike \cite{cudfkd}, an adversarial strategy is not adopted. Instead, we only use the function $\alpha(t)$ to control the difficulty of the pseudo-graphs generated by the teacher model. We set $\alpha(t)$ to a lower value at the beginning, ensuring simpler pseudo-graphs is generated in the initial stages, in order to facilitate the student model's early learning. As training progresses, $\alpha(t)$ gradually increases, allowing the student model to encounter and learn more complex graph structures. For convenience, we define $\alpha(t)$ as a piecewise function:
\begin{equation}
    \alpha(t)=\left\{\begin{aligned}
    0 \quad & t \leq k_{begin} B \\
    \alpha \cdot t  \quad & k_{begin} B < t \leq k_{end} B \\
    \lambda_{final}  \quad & t > k_{end} B
    \end{aligned}\right. \tag{7}
\label{eq7}
\end{equation}
where \(B\) denotes the number of batches of generated data, \(t\) is the current sampling batch, and \(\alpha\) is the linear coefficient. Generally, the period from epoch 0 to \(k_{begin}\)\(B\) point is considered the warm-up phase of the model. During the phase from \(k_{begin}\)\(B\) to \(k_{end}\)\(B\), the value of $\alpha(t)$ increases linearly until it stops increasing at \(k_{end}\)\(B\). 


We use a dynamic vector to control total loss variation during student model training. We compare the three strategies mentioned in \cite{cudfkd} and ultimately chose the logarithmic strategy. As it provides a smooth, continuous adjustment mechanism, enabling nuanced control over pseudo-graphs difficulty:
\begin{equation}
    \mathbf{v}^{*}(\mu, \mathcal{L})=\frac{1+e^{-\mu}}{1+e^{\mathcal{L}-\mu}}\tag{8}
\label{eq8}
\end{equation}
where \(\mu\) increases linearly with the number of training epochs, and \( \mathcal{L}\) represents the total loss during the training process.

During backpropagation, the total loss is multiplied by \(\mathbf{v}^{*}(\mu, \mathcal{L})\). Notably, the derivative of equation (\ref{eq8}) with respect to \(\mu\), i.e., \(\frac{e^{-\mu}\left(e^{\mathcal{L}}-1\right)}{\left(1+e^{\mathcal{L}-\mu}\right)^{2}}\), is always greater than 0. Consequently, \(\mathbf{v}^{*}(\mu, \mathcal{L})\) increases relative to the unit loss with each training iteration. Specifically,  \(\mathbf{v}^{*}(\mu, \mathcal{L})\) reduces the impact of smaller local loss vectors in the early stages, causing the student to pay less attention to details. However, at later stages of training, these smaller local loss vectors become more prominent, allowing the student who has already mastered most of the knowledge to start focusing on finer details.

\vspace{-6pt}
\subsection{Learnable Temperature in Adversarial Distillation}
In previous graph knowledge distillation tasks, the temperature is typically fixed, which limits the model's adaptability at different training stages. To further ensure that the student progresses from easy to hard, we incorporate an adversarially-based learnable temperature module \(\theta_{temp}\) alongside the pipeline. This module is optimized in the opposite direction of the student. The gradient is reversed to maximize the distillation loss between the student model and the teacher model in this module, thereby achieving adversarial learning. In summary, the student \(\theta_{stu}\)  and temperature module \(\theta_{temp}\) play the two-player mini-max game with the following value function \(\mathcal{L}\left(\theta_{stu}, \theta_{temp}\right)\):

\begin{equation}
\begin{aligned}
    & \min_{\theta_{stu}} \max_{\theta_{temp}} L(\theta_{stu}, \theta_{temp}) \\
    &= \min_{\theta_{stu}} \max_{\theta_{temp}} \sum_{x \in D} \alpha_1 \mathcal{L}_{cls} \left( f^s(x; \theta_{stu}), y \right)  \\
    & \quad + \alpha_2 \mathcal{L}_{div} \left( f^t(x; \theta_{tea}), f^s(x; \theta_{stu}), \theta_{temp} \right) 
\end{aligned}
\tag{9}
\label{eq9}
\end{equation}

The practical implementation of equation (\ref{eq9}) involves alternately updating the student model and the temperature module. After the student model updates its parameters, the temperature module is subsequently updated in the opposite direction. The primary parameter being updated in the temperature module is the temperature itself. This dynamically changing temperature is utilized in each epoch's distillation process by computing the KL divergence between the student and teacher outputs, as well as the loss between the student output and the true labels, thereby adaptively adjusting the training loss:
\begin{equation}
    \hat{\theta}_{stu}=\arg \min _{\theta_{stu}} \mathcal{L}\left(\theta_{ stu}, \hat{\theta}_{temp}\right) \tag{10}
\label{eq10}
\end{equation}
\begin{equation}
\hat{\theta}_{temp}=\arg \max _{\theta_{temp}} \mathcal{L}\left(\hat{\theta}_{stu}, \theta_{temp}\right) \tag{11}
\label{eq11}
\end{equation}

Equations (\ref{eq10}) and (\ref{eq11}) are implemented using the stochastic gradient descent method. Student model parameters \(\theta_{stu}\) and temperature module parameters \(\theta_{temp}\) are updated as follows:
\begin{equation}
    \theta_{stu} \leftarrow \theta_{stu}-\mu \frac{\partial \mathcal{L}}{\partial \theta_{stu}} \tag{12}
\label{eq12}
\end{equation}
\begin{equation}
    \theta_{temp} \leftarrow \theta_{temp}+\mu \frac{\partial \mathcal{L}}{\partial \theta_{temp}} \tag{13}
\label{eq13}
\end{equation}
where \(\mu\) is the learning rate.

It is important to note that to implement equation (\ref{eq13}), we employ a non-parametric gradient reversal layer. This layer is placed before the learnable temperature module, as shown in Fig. \ref{pipeline}. In the gradient reversal layer, forward propagation remains unchanged, but during backpropagation, the gradient is multiplied by a negative value \(\beta\) controlled by cosine decay, which allows for smooth, gradual adjustments that achieve stable gradient reversal:

\begin{equation}
\begin{aligned}
\beta &= \left( 1 + \cos\left( \frac{i \pi}{\text{num\_loops}} \right) \right) / 2 \\
&\quad \times (\text{max\_value} - \text{min\_value}) + \text{min\_value}
\end{aligned}
\tag{14}
\label{eq14}
\end{equation}
where \(\text{max\_value}\) and \(\text{min\_value}\) are the maximum and minimum bounds of \(\beta\).
\(\text{num\_loops}\) is the maximum number of iterations which is set manually, and \(i\) is the current epoch.

\vspace{-6pt}
\subsection{Loss Function}
The loss function \( \mathcal{L}_{N, S, \xi}\) used during the teacher model's pseudo-graphs generation can be specifically expanded as:
\begin{equation}
  \mathcal{L}_{N, S, \xi}=\alpha(t)( \mathcal{L}_{out} + \mathcal{L}_{ distr} +\mathcal{L}_{onehot}) \tag{15}
\label{eq15}
\end{equation}
where \(\mathcal{L}_{out}\) measures the difference between the teacher model’s output and the initial labels using the cross-entropy loss function. \(\mathcal{L}_{distr}\) calculates the feature distribution loss across all batch normalization layers, and \(\mathcal{L}_{onehot}\) represents the regularization loss from one-hot encoding. 

The loss function of the student model during the training phase is formulated as follows:
\begin{equation}
  \mathcal{L}_{train}=\mathbf{v}^{*}(\mu, \mathcal{L})(\mathcal{L}_{cls}+ \alpha_{div}\mathcal{L}_{div}+\alpha_{mse}\mathcal{L}_{mse}) \tag{16}
\label{eq16}
\end{equation}
where \(\mathcal{L}_{cls}\) represents the loss between the student model's output and the true labels, \(\mathcal{L}_{div}\) denotes the KL divergence loss incorporating the variable temperature. These two terms are the same as those in equation (\ref{eq9}). \(\mathcal{L}_{mse}\) is used to correct the loss caused by the dimensional changes after projecting the intermediate output in the student model.

\vspace{-6pt}
\subsection{Analysis of Computational Complexity}

\begin{table*}[h]
\caption{Test accuracies (\%) on MUTAG, PTC, and PROTEINS.}
\centering
\scriptsize{
\resizebox{\linewidth}{!}{
        \setlength\tabcolsep{1pt}
        \renewcommand\arraystretch{1.5}
\begin{tabular}{cc||ccccIccccIcccc} 
\hline\thickhline
\rowcolor{Periwinkle!20} 
Datasets  &  & \multicolumn{4}{cI}{MUTAG}&\multicolumn{4}{cI}{PTC}&\multicolumn{4}{c}{PROTEINS}\\
 \cline{3-14}
\rowcolor{Periwinkle!20}

\hline\hline
Teacher
& & GCN-5-64 & GIN-5-64 & GCN-5-64 & GIN-5-64 & GCN-5-64 & GIN-5-64 & GCN-5-64 & GIN-5-64  & GCN-5-64 & GIN-5-64 & GCN-5-64 & GIN-5-64 \\
\rowcolor{Periwinkle!20}
Student
& & GCN-3-32  & GIN-3-32  & GIN-3-32 & GCN-3-32  & GCN-3-32  & GIN-3-32  & GIN-3-32 & GCN-3-32  &GCN-3-32  & GIN-3-32  & GIN-3-32 & GCN-3-32 \\
\hline
Teacher
& 100\% training data & 89.0 & 81.2  & 89.0 &81.2  & 65.8 & 66.0 & 65.8 & 66.0 & 78.0 & 74.4 & 78.0 &74.4\\
\hline

RG    & 0 training data & 39.1$\pm$6.8 & 58.7$\pm$4.2 & 38.8$\pm$5.8 & 50.4$\pm$6.0 & 43.2$\pm$4.4 & 53.2$\pm$5.8 & 42.9$\pm$8.0 & 52.1$\pm$7.8  & 56.6$\pm$4.8 & 33.3$\pm$7.4 & 43.4$\pm$9.5 & 35.6$\pm$8.6 \\

DG   & 0 training data  & 58.6$\pm$5.1  & 59.6$\pm$2.9 &35.4$\pm$2.2 & 38.8$\pm$4.1 & 52.9$\pm$7.8 & 45.7$\pm$4.1 & 43.9$\pm$9.2 & 40.3$\pm$6.5  & 65.4$\pm$2.3 & 52.7$\pm$5.7 & 47.4$\pm$9.9 & 46.9$\pm$8.4 \\

GFKD     & 0 training data  & 70.8$\pm$4.8  & 73.2$\pm$4.2 & \underline{70.2$\pm$5.0} & 68.4$\pm$3.7 & 57.4$\pm$2.2 & 57.5$\pm$3.9 & 54.1$\pm$6.4 & 41.3$\pm$6.2 & \underline{74.7$\pm$5.5} & 60.4$\pm$4.2 & 65.5$\pm$5.1 & \underline{73.0$\pm$4.4}  \\

GFAD     & 0 training data  & \underline{73.6$\pm$5.1}  & \underline{75.7$\pm$5.8} & 69.8$\pm$4.2 & \underline{70.1$\pm$5.9} & \underline{60.4$\pm$2.9} & \underline{61.0$\pm$3.1} & \underline{56.4$\pm$3.7} & \underline{57.3$\pm$2.9} & 70.2$\pm$3.4 & \underline{70.0$\pm$4.2} & \underline{66.8$\pm$3.9} &67.2$\pm$5.0  \\

\hline

\textbf{(ours)}
& 0 training data & \textbf{88.4$\pm$5.9} & \textbf{76.8$\pm$6.8} & \textbf{85.6$\pm$4.7} & \textbf{72.6$\pm$5.4} & \textbf{65.2$\pm$4.2} & \textbf{65.8$\pm$3.3} & \textbf{64.4$\pm$4.1} & \textbf{63.1$\pm$5.5}  & \textbf{77.8$\pm$6.1} & \textbf{77.8$\pm$3.3} & \textbf{79.5$\pm$3.8} & \textbf{77.5$\pm$1.6}    \\ \bottomrule
\end{tabular}}}
\vspace{-3pt}
\label{bio}
\end{table*}

\begin{table*}[h]
\caption{Test accuracies (\%) on IMDB-B, COLLAB, and REDDIT-B.}
\centering
\scriptsize{
\resizebox{\linewidth}{!}{
        \setlength\tabcolsep{1pt}
        \renewcommand\arraystretch{1.5}
\begin{tabular}{cc||ccccIccccIcccc} 
\hline\thickhline
\rowcolor{Periwinkle!20} 
  Datasets  &  & \multicolumn{4}{cI}{IMDB-B}&\multicolumn{4}{cI}{COLLAB}&\multicolumn{4}{c}{REDDIT-B}\\
 \cline{3-14}
\rowcolor{Periwinkle!20}
\hline\hline
Teacher 
& & GCN-5-64 & GIN-5-64 & GCN-5-64 & GIN-5-64 & GCN-5-64 & GIN-5-64 & GCN-5-64 & GIN-5-64  & GCN-5-64 & GIN-5-64 & GCN-5-64 & GIN-5-64 \\
\rowcolor{Periwinkle!20}
Student
& & GCN-3-32  & GIN-3-32  & GIN-3-32 & GCN-3-32  & GCN-3-32  & GIN-3-32  & GIN-3-32 & GCN-3-32  &GCN-3-32  & GIN-3-32  & GIN-3-32 & GCN-3-32 \\
\hline
Teacher 
&100\% training data & 73.0  & 73.5  &73.0 & 73.5   & 73.2 & 69.3 & 73.2 & 69.3 & 81.2 &75.3 & 81.2 &75.3\\
\hline

RG/DG   & 0 training data & 58.5$\pm$3.7 & 58.7$\pm$4.2 & 55.4$\pm$3.4 & 56.8$\pm$2.9 & 34.8$\pm$9.0 &28.4$\pm$7.3 & 27.2$\pm$6.3 & 30.5$\pm$7.6 & 50.1$\pm$1.0 & 49.9$\pm$0.8 & 48.9$\pm$2.1 & 47.2$\pm$3.2 \\

GFKD     & 0 training data  & 62.0$\pm$3.1 & 67.8$\pm$3.8 & \underline{67.1$\pm$2.3} &62.3$\pm$2.6 & \underline{67.3$\pm$2.4} & \textbf{65.4$\pm$2.7} & \underline{60.4$\pm$2.8} & \textbf{60.1$\pm$3.4} & 66.5$\pm$3.7 & 63.8$\pm$4.5 & 63.1$\pm$5.7 & 67.8$\pm$3.5  \\

GFAD     & 0 training data  & \underline{67.8$\pm$3.9}  & \underline{70.1$\pm$4.3} & 66.5$\pm$4.3 & \underline{65.2$\pm$5.0} & 62.5$\pm$3.2 & \underline{65.1$\pm$2.7} & 59.8$\pm$4.8 & \underline{56.5$\pm$4.0} & \underline{68.7$\pm$2.8} & \underline{68.4$\pm$2.4} & \underline{67.0$\pm$2.7} & \underline{67.9$\pm$5.3}  \\

\hline

\textbf{(ours)}
& 0 training data & \textbf{69.1$\pm$2.2} & \textbf{70.9$\pm$2.3} & \textbf{68.1$\pm$2.9} & \textbf{66.1$\pm$2.5} & \textbf{67.7$\pm$3.3} & 61.1$\pm$2.9 & \textbf{60.8$\pm$3.1} & 52.0$\pm$2.8 & \textbf{75.7$\pm$2.7} & \textbf{73.1$\pm$3.2} & \textbf{73.1$\pm$3.4} & \textbf{69.7$\pm$2.4}   \\ \bottomrule
\end{tabular}}}
\vspace{-5pt}
\label{social}
\end{table*}

We conduct a detailed comparison of the computational complexity between the Bernoulli-based sampling method and ours to validate the justification of our motivation. Assuming a total of \(k\) batches of pseudo-graphs need to be generated, \(n\) is the average number of nodes per graph, \(\xi\) is the average number of nodes reduced per graph during optimization and \(\text{num\_loops}\) represents the total number of iterations. The following analysis focuses on each batch individually.

\vspace{0.5em}  
\noindent\textbf{\small Comparison of Forward Propagation}

In the Bernoulli sampling method, the feature and structure parameters are optimized separately. The feature parameters of nodes require a single forward computation, while the structure parameters need two additional noise terms, each followed by a forward computation to obtain gradients from the difference in losses. Thus, a total of three forward propagations are needed.

Our method updates the gradients of both feature and structure parameters simultaneously, requiring only one forward computation.

\vspace{0.5em}  
\noindent\textbf{\small Comparison of Noise Sampling}
\vspace{0.25em}    

The Bernoulli-based method performs two additional full-graph noise samplings for each data batch, and since this operation is required in every iteration, it must be multiplied by \(\text{num\_loops}\). Therefore, the time complexity for generating noise is \(\text{num\_loops} \cdot O\left(2 \cdot n^{2}\right)\).

In contrast, our method only requires noise addition during initialization, leaving the remaining process to be handled automatically by gradient optimization. Thus, with the introduction of \(\xi\), our time complexity is \(O\left( (n-\xi)^{2}\right)\).

\vspace{0.5em}  
\noindent\textbf{\small Comparison of Gradient Computation}
\vspace{0.25em}    

The Bernoulli-based method requires a loop algorithm to manually update structural parameter's gradient in every iteration, which also increases memory overhead.

Our method, however, simplifies the computation process by leveraging PyTorch's automatic differentiation framework to directly track the computation graph, enabling efficient gradient calculation.

Each of the above comparative effects should be further multiplied by \(k\). These results clearly demonstrate that our method significantly accelerates both pseudo-graphs generation and the distillation process.

\section{Experiments}
\label{experiment}
In this section, we present comprehensive experiments to evaluate ACGKD. It is important to emphasize that our objective is not to generate realistic graphs, but rather to maximize the knowledge transfer from a pretrained teacher GNN to a student GNN without relying on any training data.
\vspace{-8pt}
\subsection{Detailed settings of the experiment}
We utilize six graph classification benchmark datasets from \cite{r1} to pretrain the teacher models, consisting of three bioinformatics graph datasets (MUTAG, PTC, and PROTEINS) and three social network graph datasets (IMDB-B, COLLAB, and REDDIT-B). The dataset statistics are provided in Table \ref{dataset-table}. For each dataset, 70\% of the data is used to pretrain the teacher models, while the remaining 30\% is reserved for testing.

We use two well known GNN architectures, i.e, GCN \cite{r3} and GIN \cite{r1}. We assign these two architectures to the teacher and student models, where the teacher and student can be either GCN or GIN, resulting in 2×2 combinations. We use the form of (architecture-layer number-feature dimensions) to denote a GNN. For example, GIN-5-64 represents a GNN with 5 GIN layers and 64 feature dimensions.

We use GFKD \cite{gfkd}, GFAD \cite{gfad} and the two models GFKD defines as baselines for data-free methods, followed by extensive experiments to verify the superiority of ACGKD:
\begin{itemize}
     \item \textbf{Random Graphs (RG)}: RG generates graphs by randomly drawing from a uniform distribution as node features and topological structures, and then use these graphs to transfer knowledge.

    \item \textbf{DeepInvG(DG)}: As the original DeepInversion \cite{yin2020dreaming} cannot learn the structures of graph data, here DG first randomly generates graph structures and then uses DeepInversion to learn node and structure features with the objective $C(Y, T(H, A)) + R_{bn}$.
    
    \item \textbf{GFKD}: GFKD facilitates knowledge transfer by learning the topological structures of graphs through modeling them as a multivariate Bernoulli distribution. 

    \item \textbf{GFAD}: GFAD uses a generator to construct pseudo-graphs and keeps the teacher model fixed. The generator and student model are alternately trained to achieve an adversarial effect.
\end{itemize}

It is important to emphasize that ACGKD does not require an additional generator like GFAD; instead, it directly uses the teacher model to generate pseudo-graphs. During the graph generation phase,We set \(\text{num\_loops}\) to 1800 to optimize the node and structure parameters. The initial learning rates for the structure and feature parameters are set to 1.0 and 0.01, respectively, with an exponential decay over time, and we select \(k_{begin}\) = 0.1 and \(k_{end}\) = 0.9. For knowledge distillation, all GNNs are trained for 400 epochs with Adam, and the learning rate is linearly decreased from 1.0 to 0. 

\begin{table*}[t]
\caption{Ablation Study on curriculum learning, dynamic temperature, classifier reuse.}
\centering
\scriptsize{
\resizebox{\linewidth}{!}{
        \setlength\tabcolsep{1pt}
        \renewcommand\arraystretch{1.5}
\begin{tabular}{cc||ccccIccccIcccc} 
\hline\thickhline
\rowcolor{Periwinkle!20} 
 Teacher  &    & GCN-5-64  & GIN-5-64  & GCN-5-64 & GIN-5-64  & GCN-5-64 & GIN-5-64 & GCN-5-64 & GIN-5-64 & GCN-5-64 & GIN-5-64 & GCN-5-64 & GIN-5-64 \\
\rowcolor{Periwinkle!20}
 Student  &  & GCN-3-32  & GIN-3-32  & GIN-3-32 & GCN-3-32  & GCN-3-32 & GIN-3-32  & GIN-3-32 & GCN-3-32 & GCN-3-32 & GIN-3-32 & GIN-3-32 & GCN-3-32 \\
 \cline{3-14}
 \hline\hline
\rowcolor{Periwinkle!20}
 Datasets  &  & \multicolumn{4}{cI}{MUTAG}&\multicolumn{4}{cI}{PTC}&\multicolumn{4}{c}{PROTEINS}\\ \hline
W/o CL   &   & \underline{88.2$\pm$1.8} & 73.0$\pm$3.8 & 84.9$\pm$2.6 &  70.2$\pm$1.4 & \underline{64.0$\pm$2.6} & \underline{64.6$\pm$1.0} & 64.2$\pm$1.7  & 60.1$\pm$3.5 & 77.4$\pm$2.2 & \underline{76.5$\pm$1.4} & \underline{79.4$\pm$2.4} & 76.6$\pm$1.7 \\
W/o DT    &  & \underline{88.2$\pm$2.1}  & 72.2$\pm$1.9  & \underline{85.5$\pm$2.4}  & 69.8$\pm$2.3  & 62.3$\pm$1.8 & 64.2$\pm$2.1 & \underline{64.4$\pm$1.5} & \underline{60.6$\pm$1.8}  & \underline{77.6$\pm$3.0} & 75.9$\pm$2.7 & 79.0$\pm$2.9 & 74.4$\pm$3.1 \\
W/o CR    &      & 80.0$\pm$2.8  & \underline{74.4$\pm$3.2} & 78.6$\pm$1.9 & \underline{71.6$\pm$2.8} & 63.5$\pm$1.2 & 64.2$\pm$3.8 & 61.7$\pm$2.1 &60.0$\pm$3.0 & 77.5$\pm$0.8 & 74.9$\pm$2.9 & 77.8$\pm$2.5 & \underline{76.9$\pm$3.2}  \\
ACGKD &  & \textbf{88.4$\pm$5.9} & \textbf{76.8$\pm$6.8} & \textbf{85.6$\pm$4.7} & \textbf{72.6$\pm$5.4} & \textbf{65.2$\pm$4.2} & \textbf{65.8$\pm$3.3} & \textbf{64.4$\pm$4.1} & \textbf{63.1$\pm$5.5}  & \textbf{77.8$\pm$6.1} & \textbf{77.8$\pm$3.3} & \textbf{79.5$\pm$3.8} & \textbf{77.5$\pm$1.6}  \\
 \hline\hline
\rowcolor{Periwinkle!20}
 Datasets  &  & \multicolumn{4}{cI}{IMDB-B}&\multicolumn{4}{cI}{COLLAB}&\multicolumn{4}{c}{REDDIT-B}\\ \hline
W/o CL   &     & \underline{68.1$\pm$1.7} & 67.5$\pm$1.3 & 66.3$\pm$2.1 & 62.9$\pm$1.2  & \underline{64.3$\pm$1.5} & 60.7$\pm$1.1 & \underline{53.0$\pm$2.2} & 40.9$\pm$3.0 & 73.5$\pm$2.9 &71.1$\pm$3.3 & 73.1$\pm$2.3 & 67.1$\pm$1.9 \\
W/o DT    &   & 66.5$\pm$1.9  & 65.9$\pm$2.1 & 68.1$\pm$3.2 &63.3$\pm$1.5 & \underline{64.3$\pm$1.7} & 59.5$\pm$3.7 & 52.4$\pm$2.8 & 40.1$\pm$3.4 & \underline{75.7$\pm$2.7} & 70.3$\pm$3.5 & 72.1$\pm$4.7 & 67.0$\pm$2.1  \\
W/o CR  &     & 65.9$\pm$3.1  & \underline{67.6$\pm$2.4} & \underline{72.5$\pm$2.7} & \underline{66.0$\pm$2.4} & 63.9$\pm$2.5 & \textbf{68.2$\pm$1.8} & 52.0$\pm$3.1 & \textbf{60.3$\pm$2.1} & 73.3$\pm$1.4 & \underline{73.0$\pm$1.9} & \underline{75.0$\pm$0.7} & \underline{70.0$\pm$3.1}  \\
ACGKD & & \textbf{69.1$\pm$2.2} & \textbf{70.9$\pm$2.3} & \textbf{68.1$\pm$2.9} & \textbf{66.1$\pm$2.5} & \textbf{67.7$\pm$3.3} & \underline{61.1$\pm$2.9} & \textbf{60.8$\pm$3.1} & \underline{52.0$\pm$2.8} & \textbf{75.7$\pm$2.7} & \textbf{73.1$\pm$3.2} & \textbf{73.1$\pm$3.4} & \textbf{69.7$\pm$2.4}   \\ \bottomrule
\end{tabular}}}
 \vspace{-3pt}
\label{ablation}
\end{table*}

\begin{table}[h!]
  \centering
  \begin{tabular}{lrrr}
    \toprule
    Dataset & \#Graphs & \#Classes & Avg\#Graph Size \\
    \midrule
    MUTAG   & 188   & 2  & 17.93 \\
    PTC     & 344   & 2  & 14.29 \\
    PROTEINS & 1,113 & 2  & 39.06 \\
    IMDB-B  & 1,000 & 2  & 19.77 \\
    COLLAB  & 5,000 & 3  & 74.49 \\
    REDDIT-B & 2,000 & 2  & 429.62 \\
    \bottomrule
  \end{tabular}
\setlength{\abovecaptionskip}{0.1cm}
\caption{Summary of datasets.}
\vspace{-15pt}
\label{dataset-table}
\end{table}

\vspace{2pt}
\begin{figure}[h!]
    \centering
    \includegraphics[width=\columnwidth]{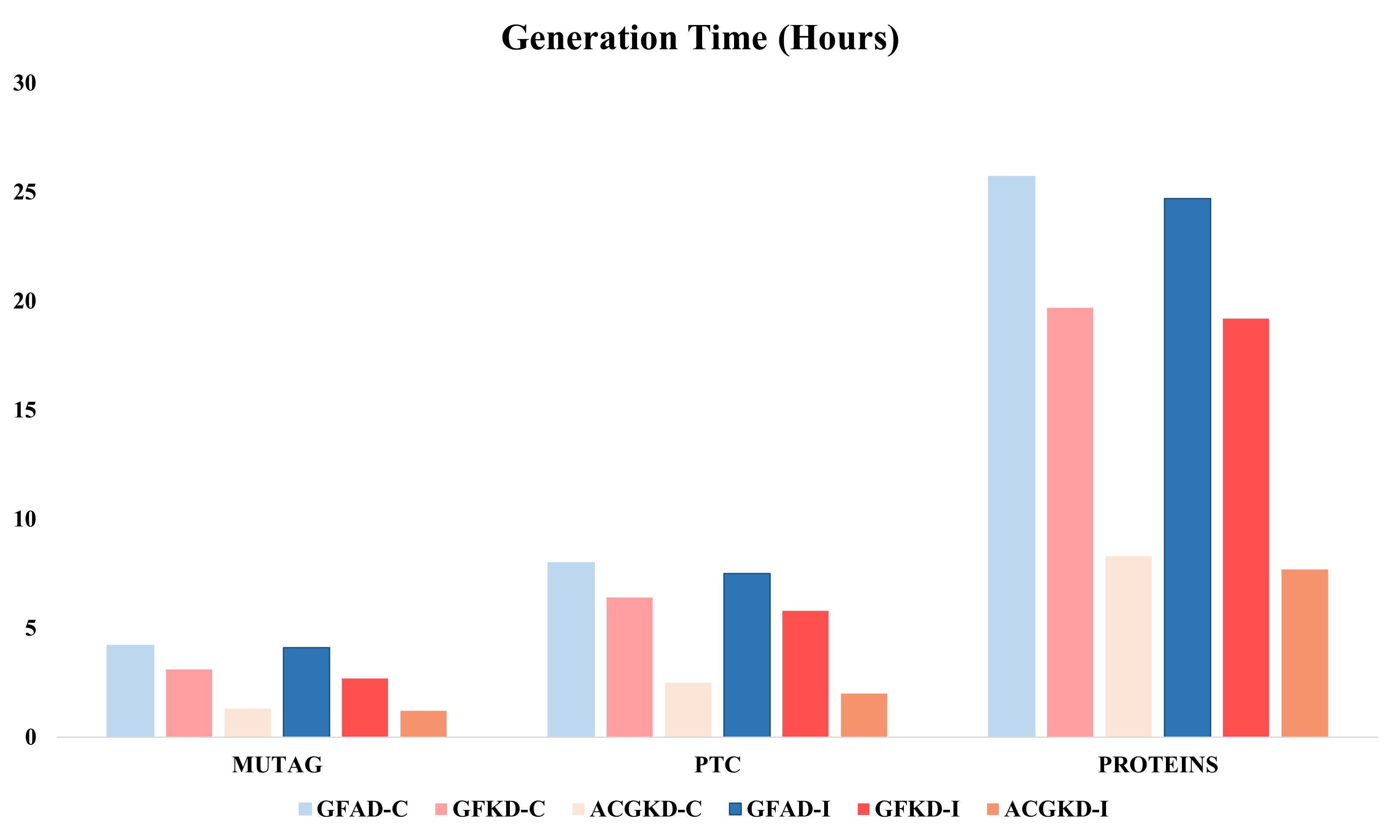}  
    \captionsetup{justification=centering}  
    \caption{\normalsize Comparison of bioinformatic datasets generation time}  
    \vspace{-10pt}
    \label{time-table1}  
\end{figure}

\begin{figure}[h!]
    \centering
    \includegraphics[width=\columnwidth]{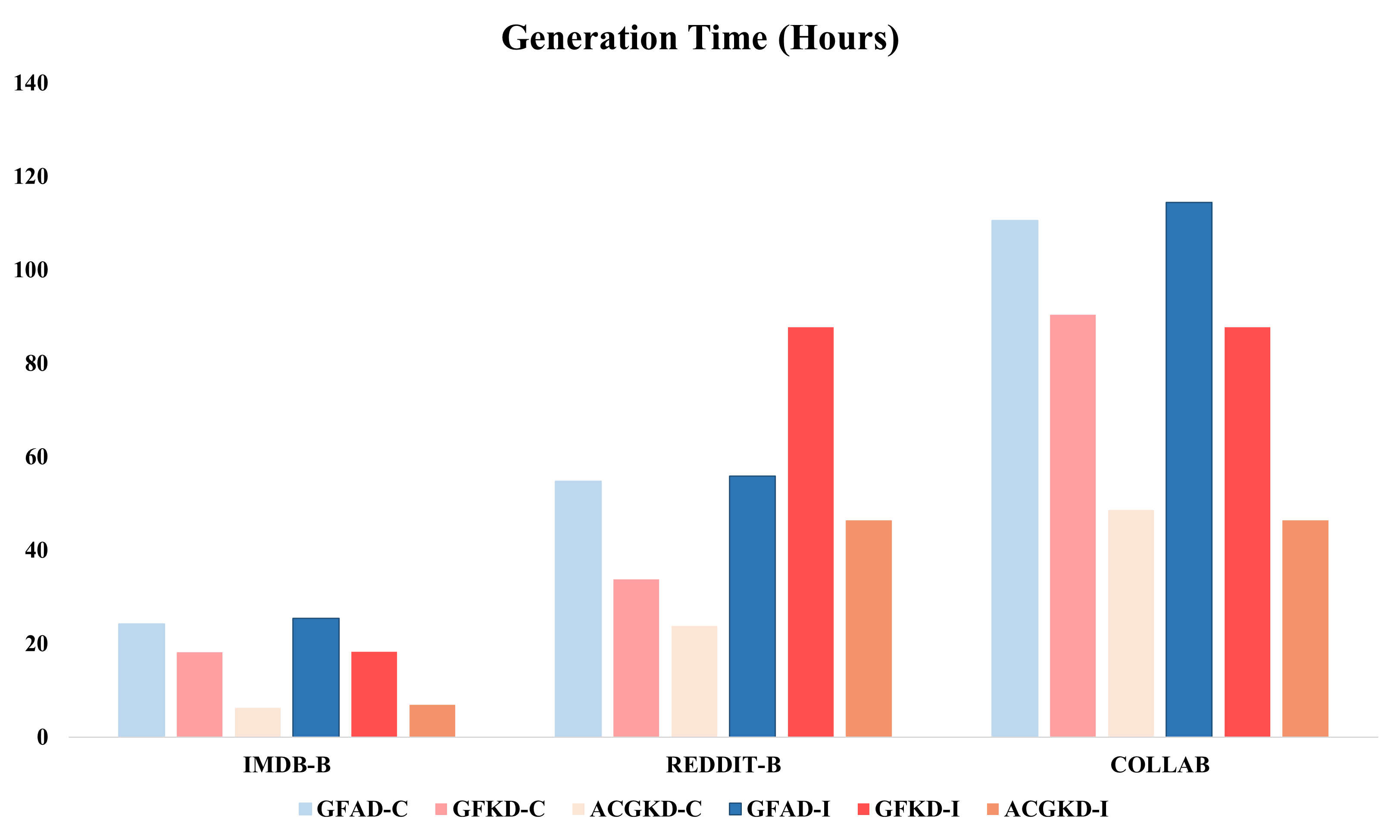}  
    \captionsetup{justification=centering}  
    \caption{\normalsize Comparison of social datasets generation time}  
    \vspace{-4pt}
    \label{time-table2}  
\end{figure}

\begin{figure}[h!]
    \centering
    \includegraphics[width=\columnwidth]{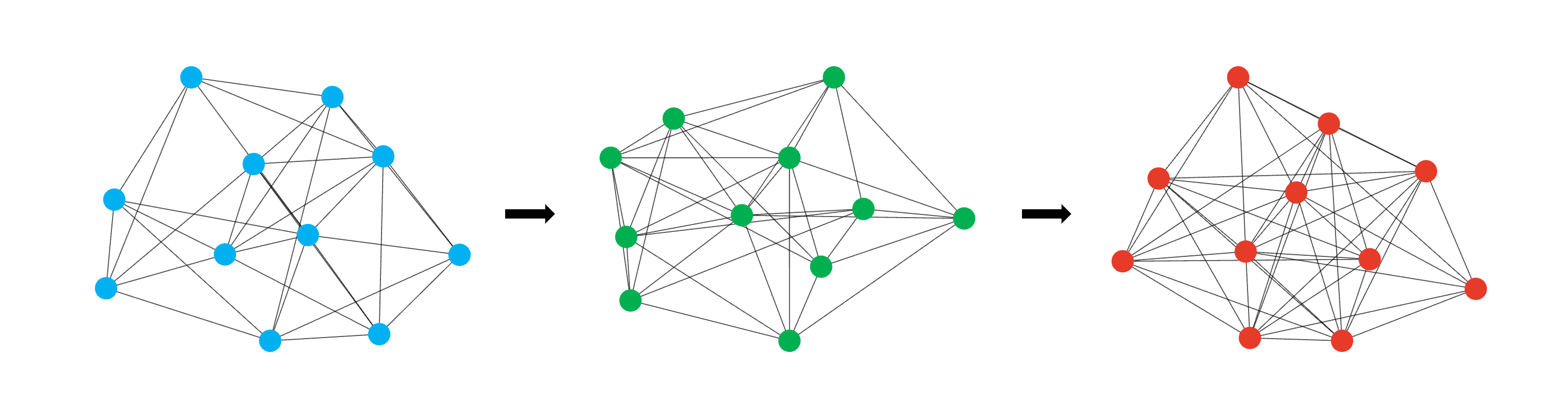}  
    \captionsetup{justification=centering}  
    \caption{\normalsize Pseudo-graphs generated by the teacher at different times under the easy-to-hard mode}  
    \label{curriculum}  
\end{figure}
\vspace{6pt}

\subsection{Results and Comparisons}
\vspace{-2pt}
\noindent\textbf{\small Experiments on Bioinformatics Graph Data}
\vspace{0.15em} 

Table \ref{bio} presents the comparison results on three bioinformatics graph datasets. As expected, the overall performance of RG is worse than the other methods, as the teacher's knowledge is poorly represented in random graphs. Both GFKD and DG learn node features for pseudo-graphs, but GFKD also learns the graph structures, whereas DG generates structures randomly. Therefore, GFKD performs slightly better. Meanwhile, the adversarial approach used by GFAD further improves performance. In contrast, our proposed ACGKD significantly improves the performance across all the three datasets. For instance, the accuracy improvement of ACGKD is 17.6\% over GFKD and 14.8\% over GFAD with teacher GCN-5-64 and student GCN-3-32 on MUTAG. This strongly demonstrates the effectiveness of our method. Moreover, we observe that ACGKD consistently outperforms GFKD and GFAD across different teacher-student architecture combinations, highlighting the generalizability of our method for Graph Neural Networks. 

\begin{figure*}[t]
    \centering
    \begin{tikzpicture}
        \node (img1) at (-1, 0) {\includegraphics[width=0.24\textwidth,height=0.18\textwidth]{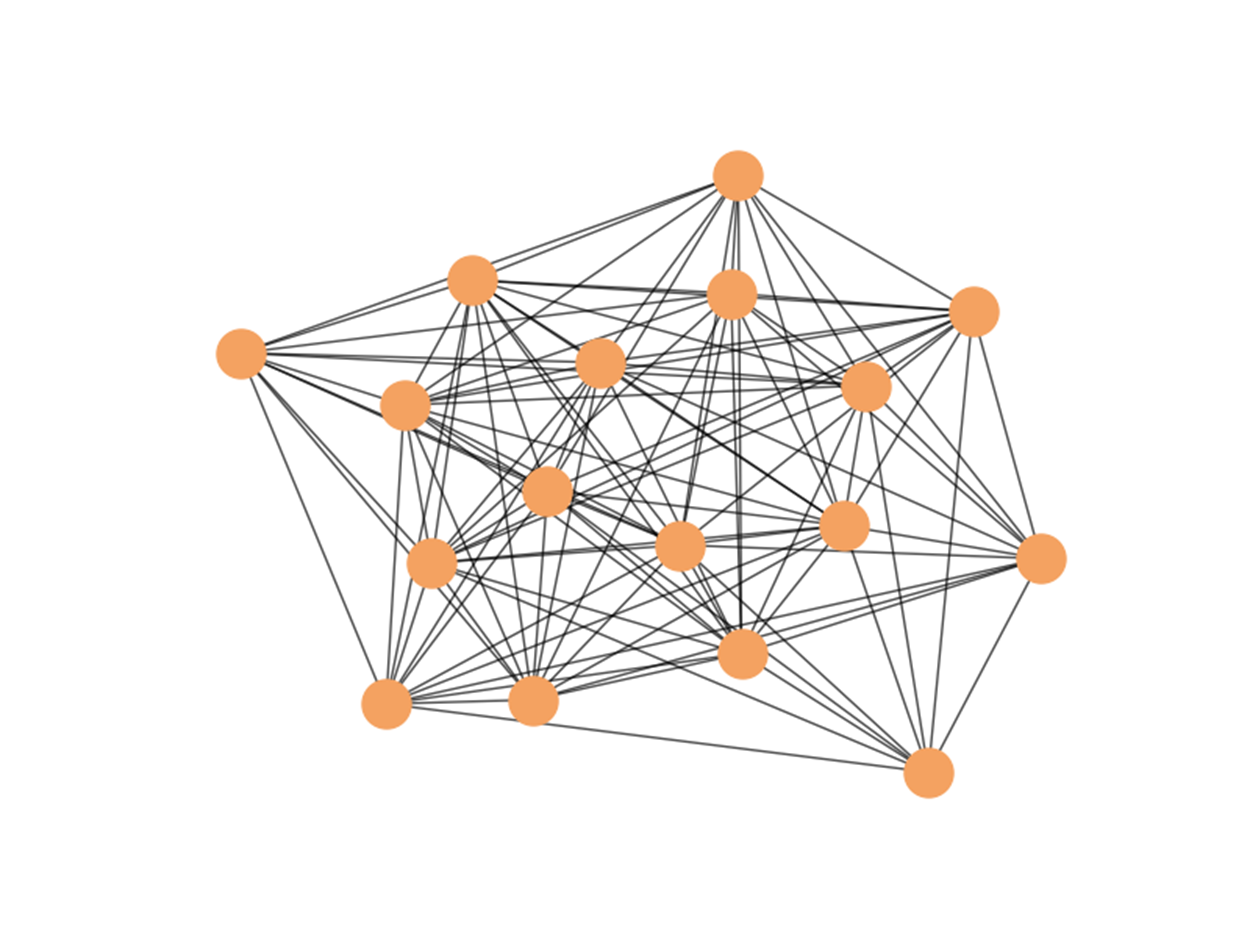}};
        \node (img2) at (3.2, 0) {\includegraphics[width=0.24\textwidth,height=0.18\textwidth]{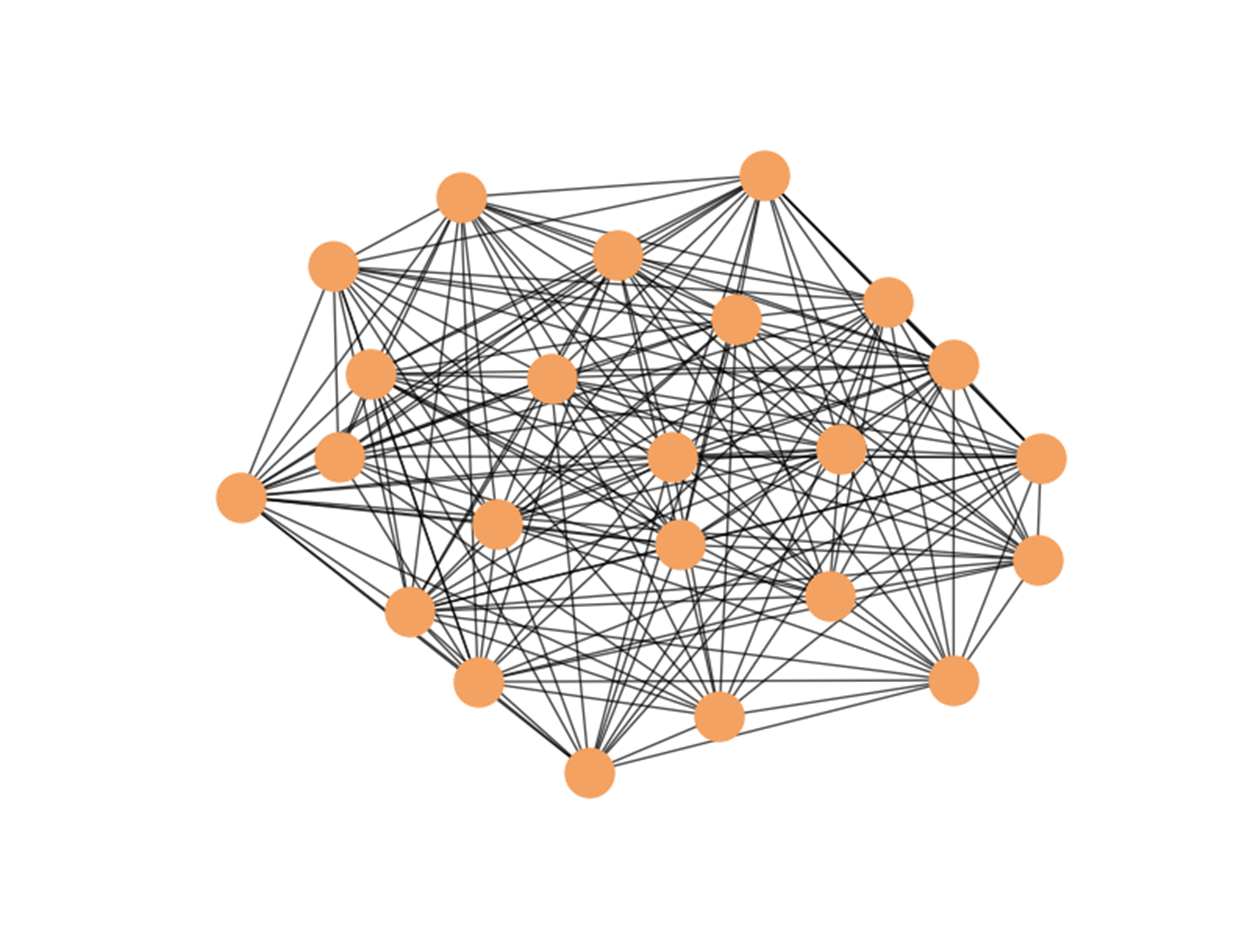}};
        \node (img3) at (7.4, 0) {\includegraphics[width=0.24\textwidth,height=0.18\textwidth]{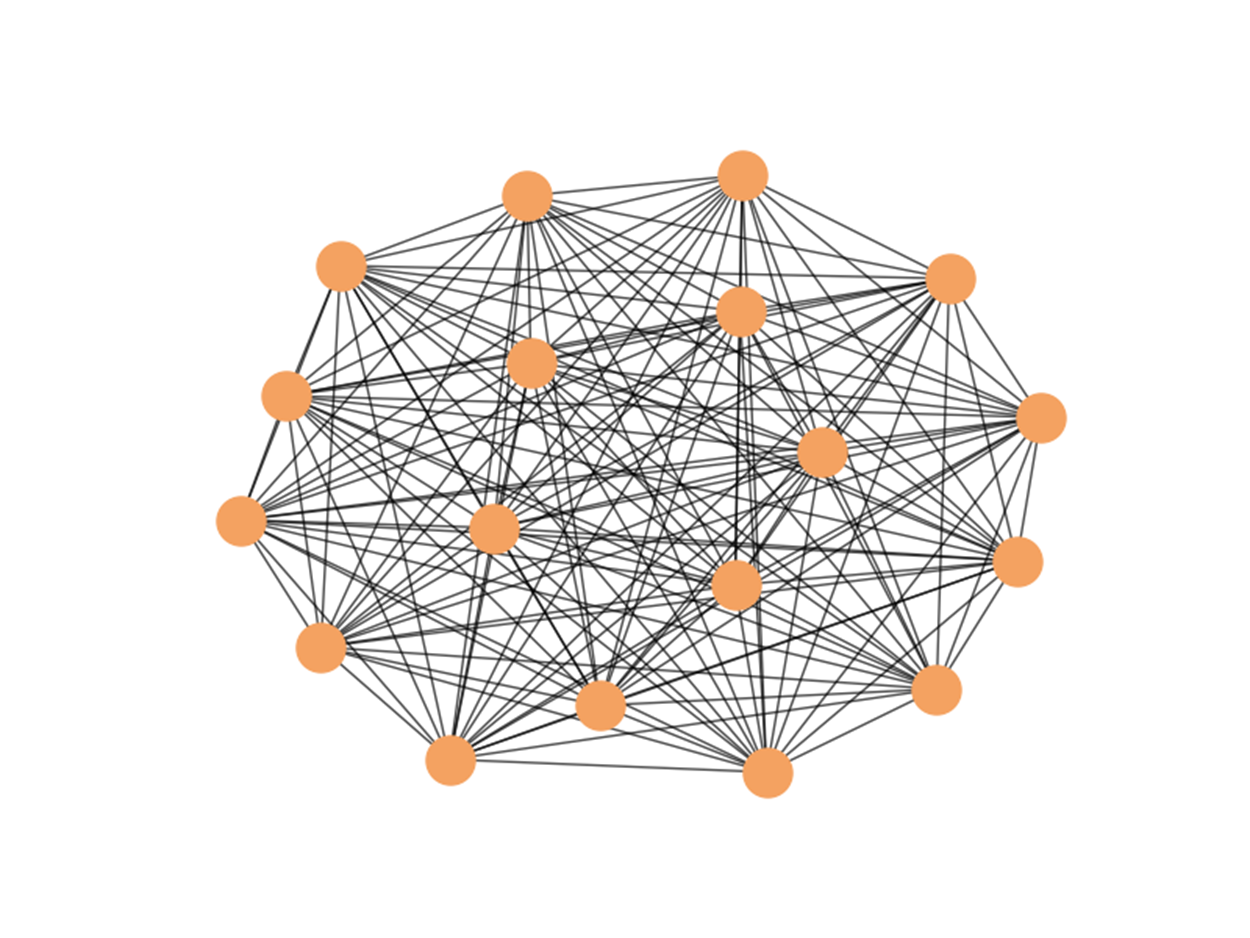}};
        \node (img4) at (11.6, 0) {\includegraphics[width=0.24\textwidth,height=0.18\textwidth]{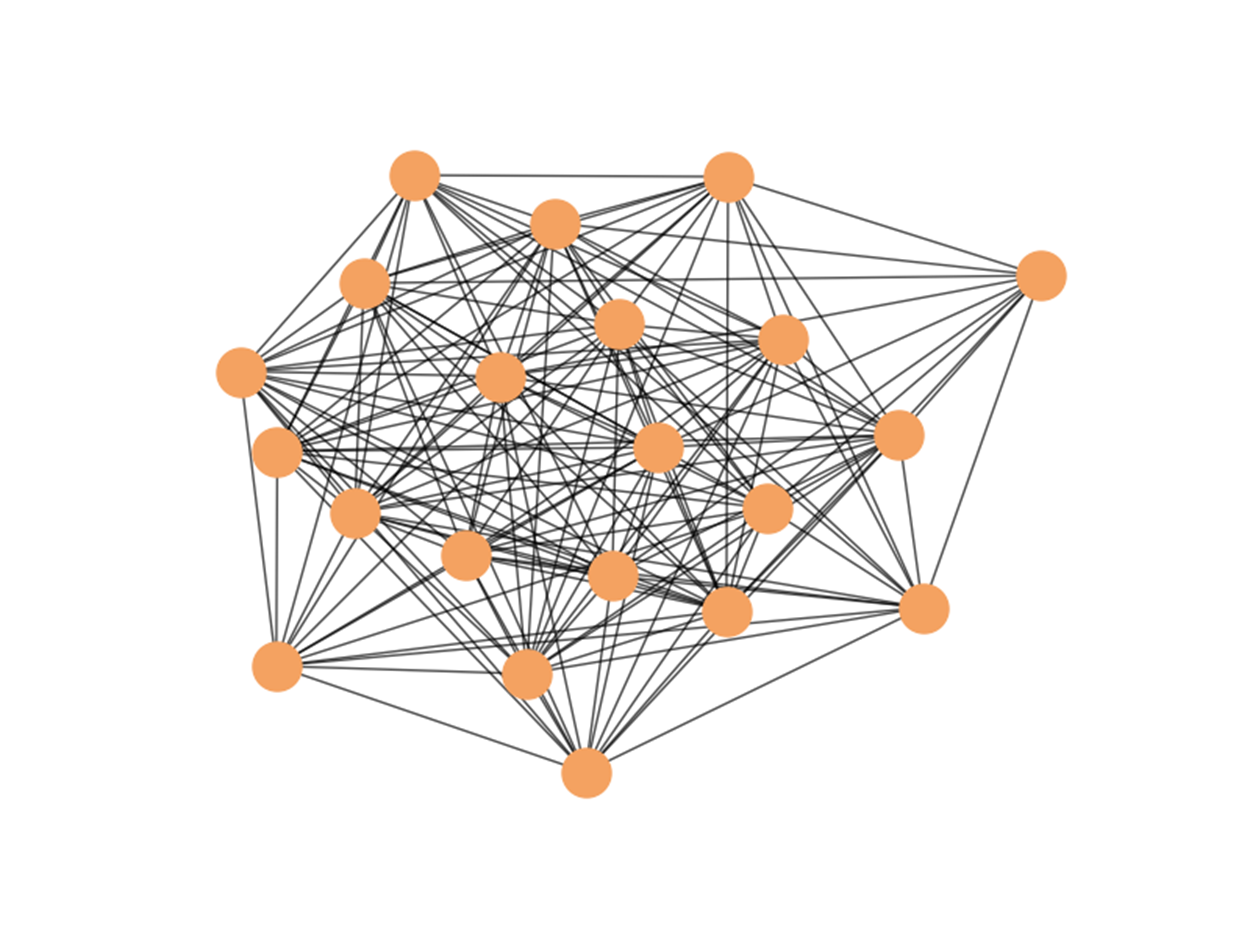}};
        \node (img5) at (-1, -3) {\includegraphics[width=0.24\textwidth,height=0.18\textwidth]{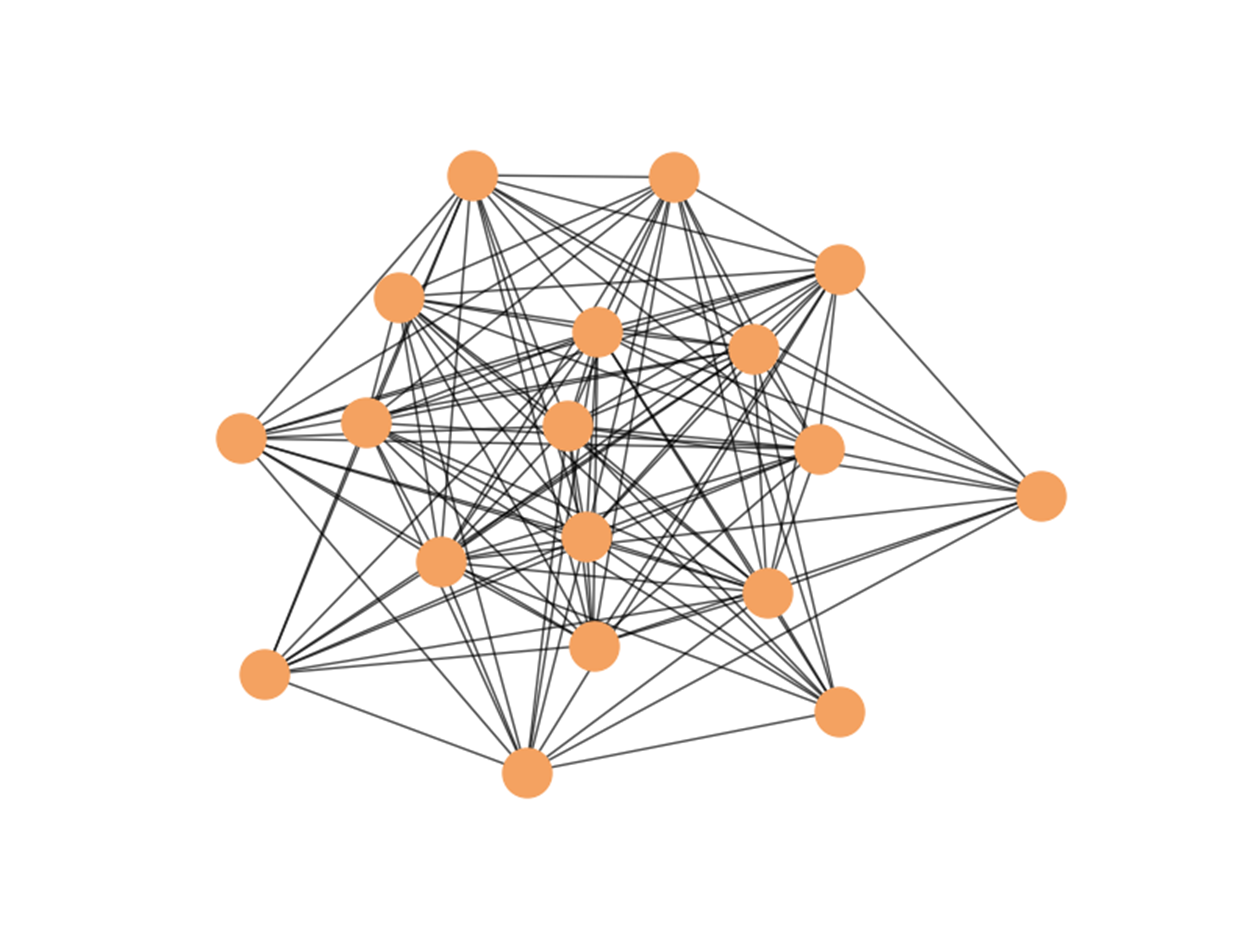}};
        \node (img6) at (3.2, -3) {\includegraphics[width=0.24\textwidth,height=0.18\textwidth]{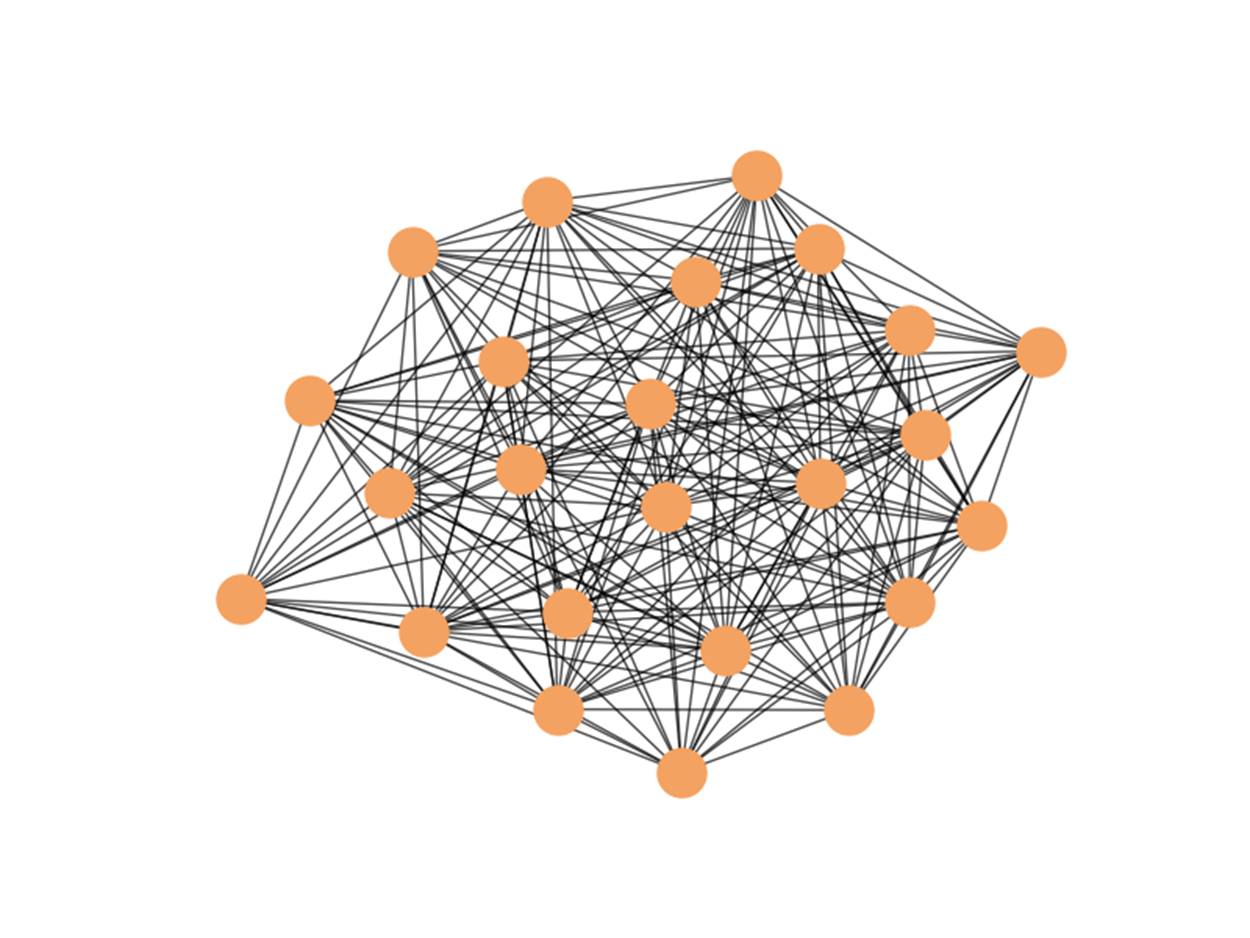}};
        \node (img7) at (7.4, -3) {\includegraphics[width=0.24\textwidth,height=0.18\textwidth]{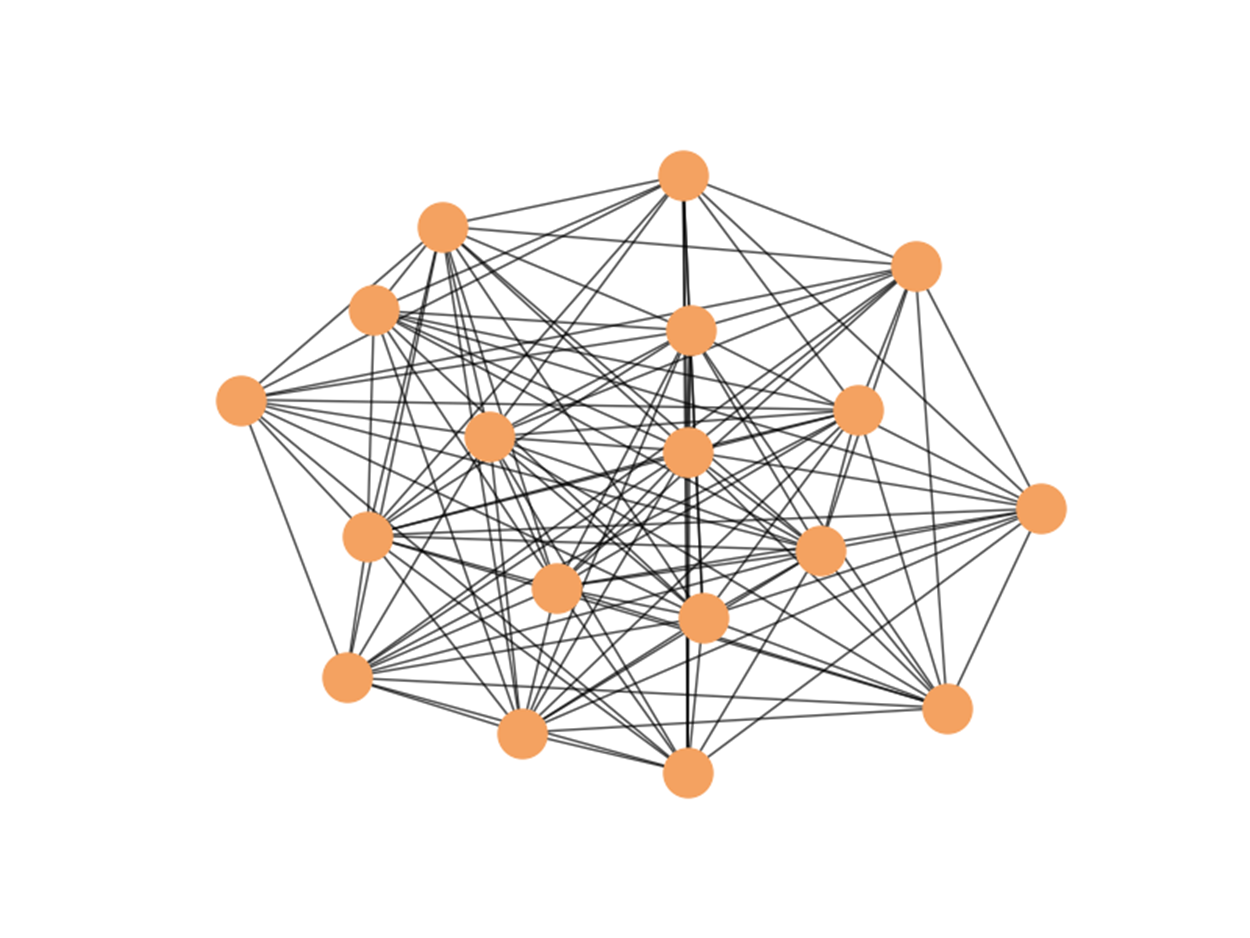}};
        \node (img8) at (11.6, -3) {\includegraphics[width=0.24\textwidth,height=0.18\textwidth]{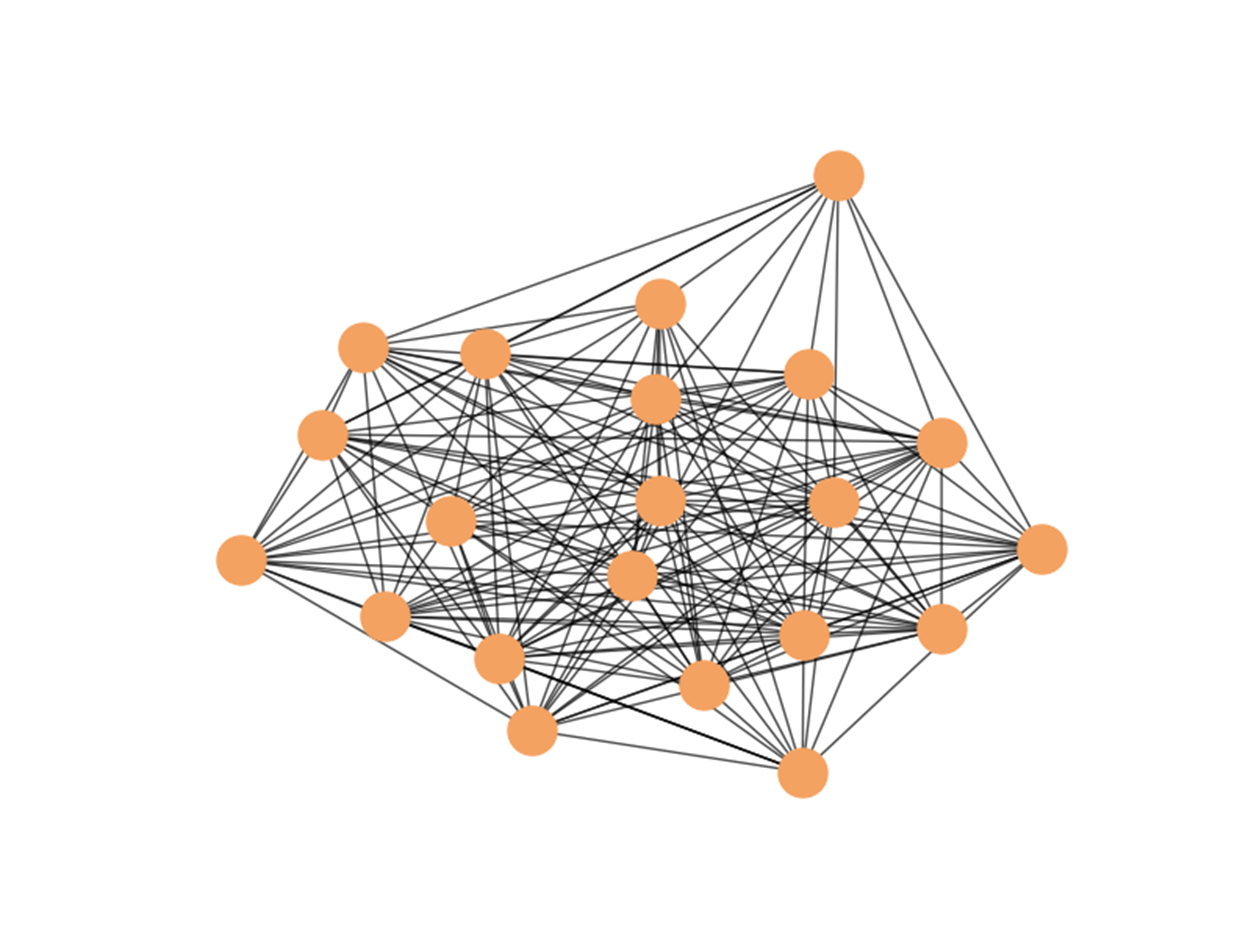}};
        \node (img9) at (-1, -6) {\includegraphics[width=0.24\textwidth,height=0.18\textwidth]{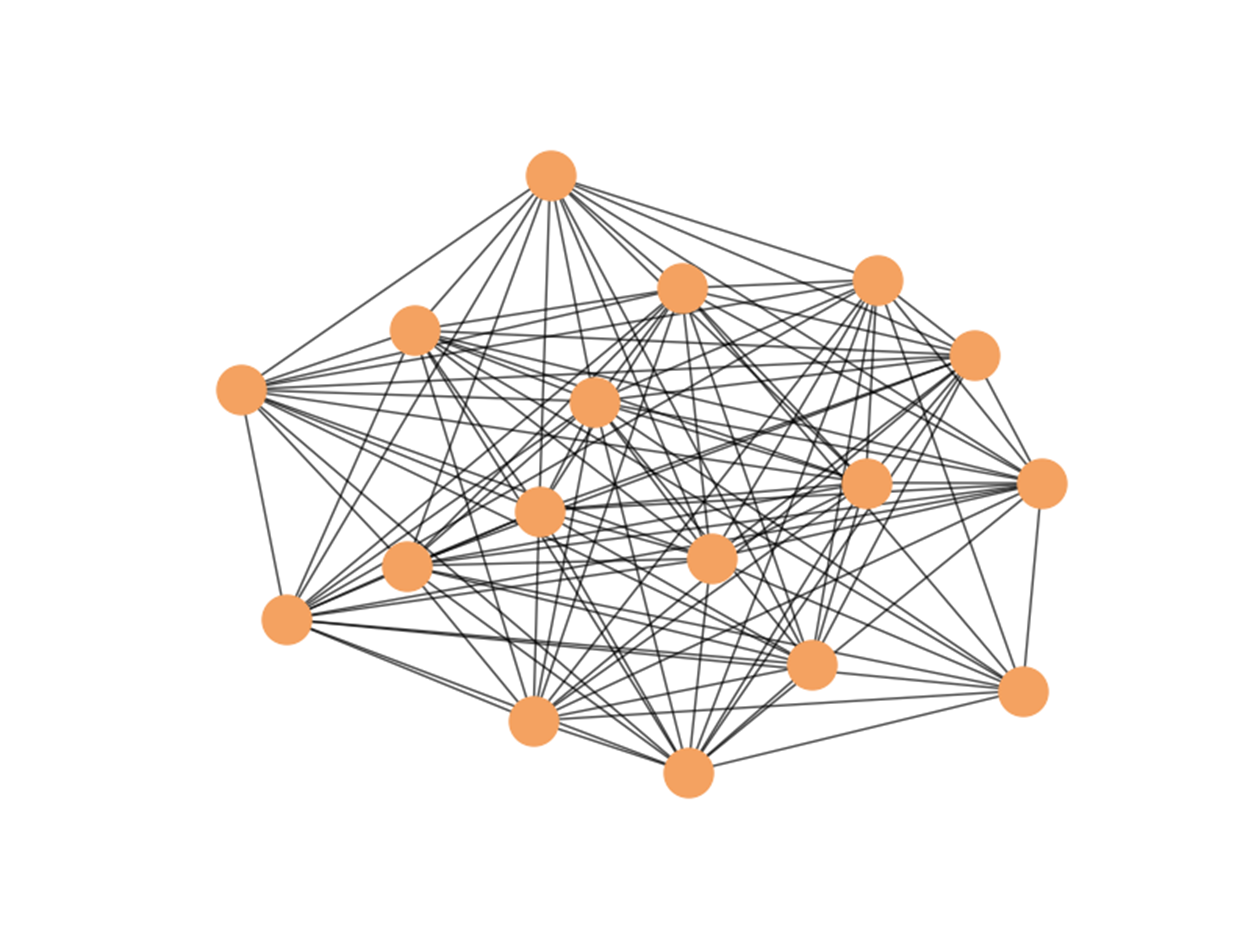}};
        \node (img10) at (3.2, -6) {\includegraphics[width=0.24\textwidth,height=0.18\textwidth]{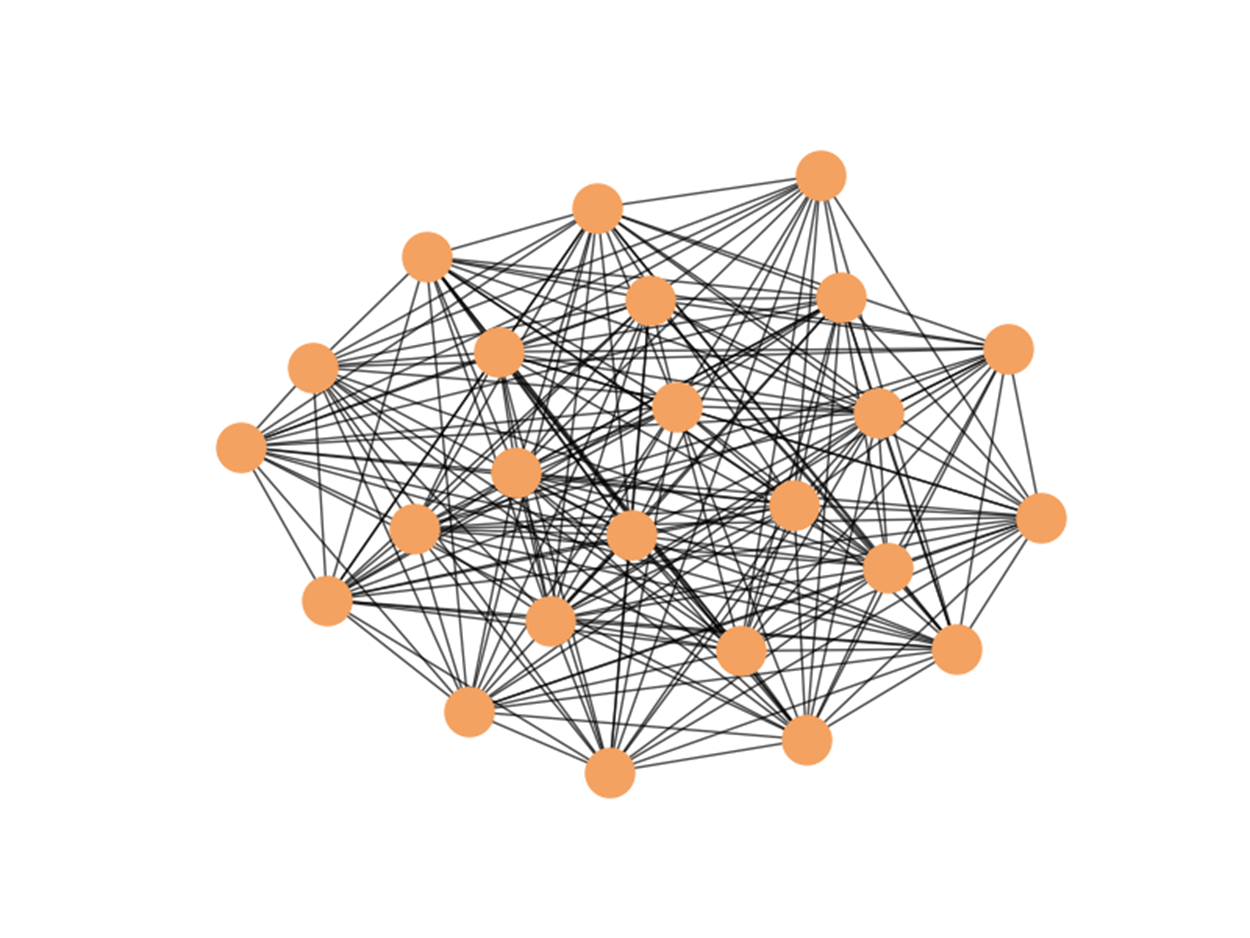}};
        \node (img11) at (7.4, -6) {\includegraphics[width=0.24\textwidth,height=0.18\textwidth]{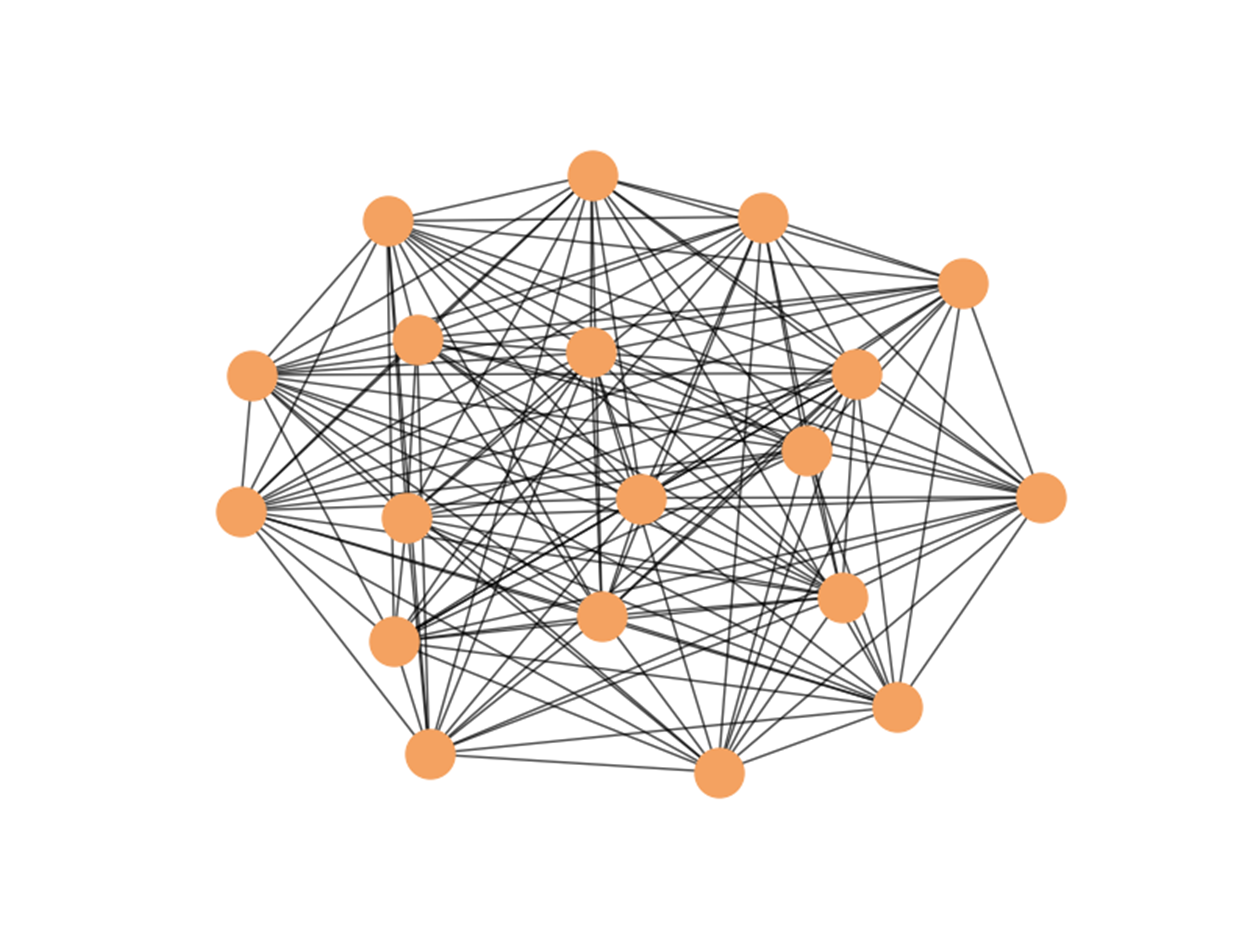}};
        \node (img12) at (11.6, -6) {\includegraphics[width=0.24\textwidth,height=0.18\textwidth]{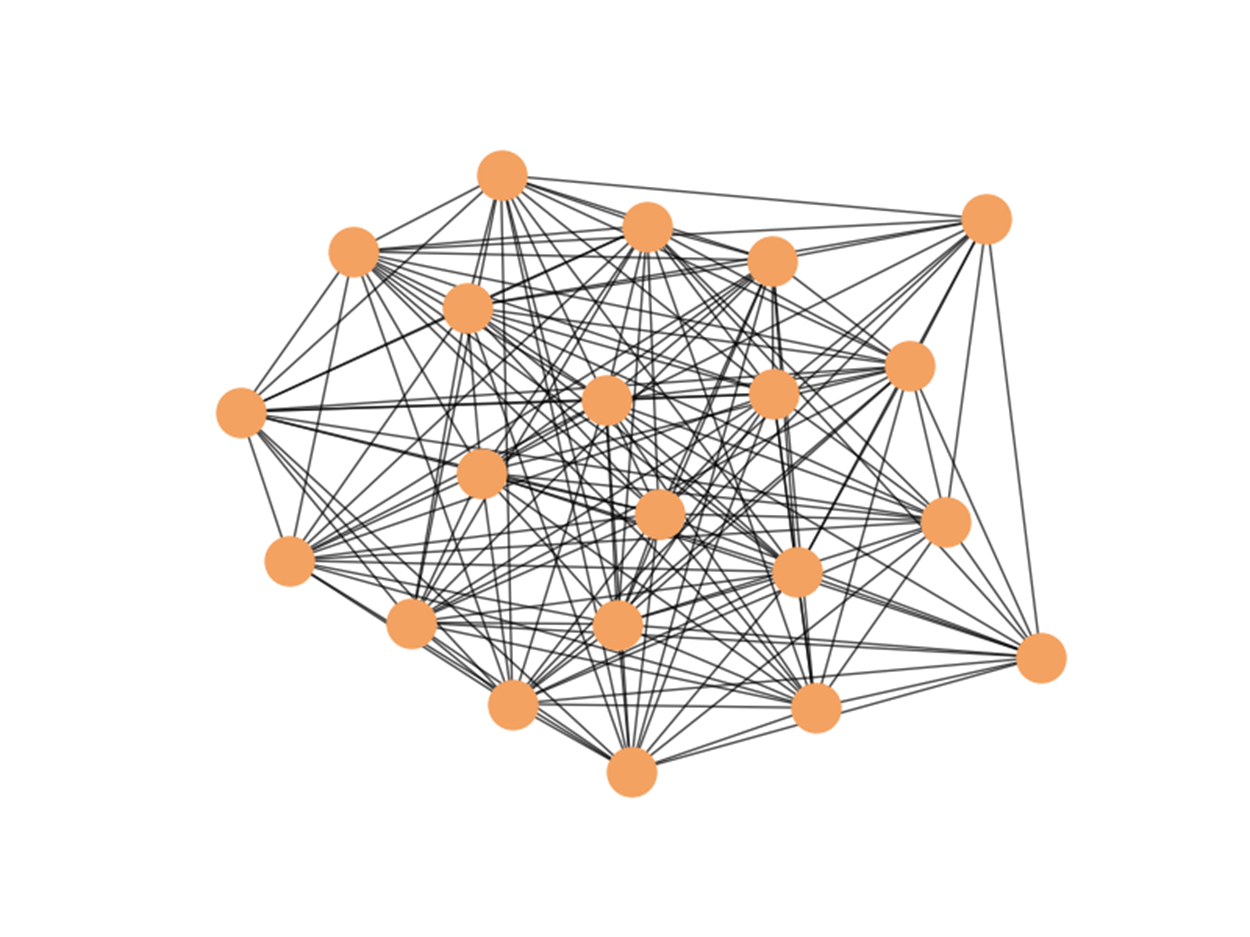}};
        \node (img13) at (-1, -9) {\includegraphics[width=0.24\textwidth,height=0.18\textwidth]{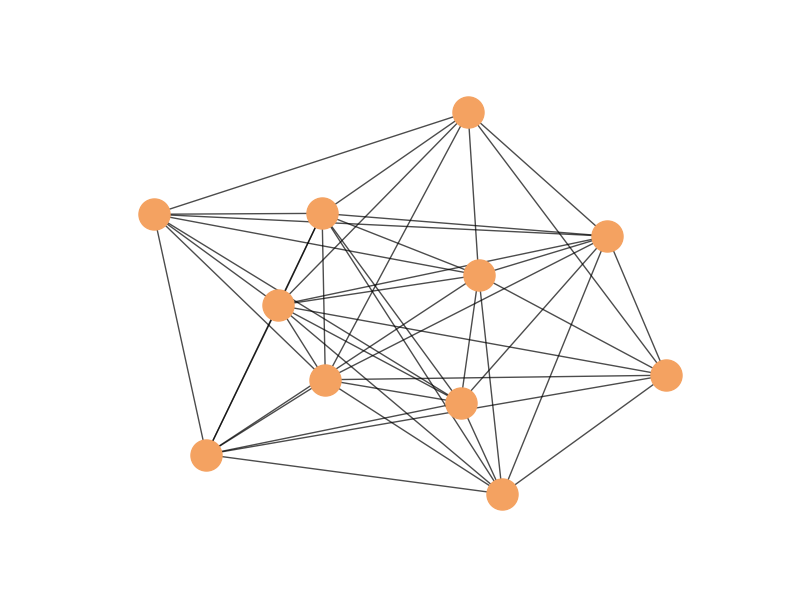}};
        \node (img14) at (3.2, -9) {\includegraphics[width=0.24\textwidth,height=0.18\textwidth]{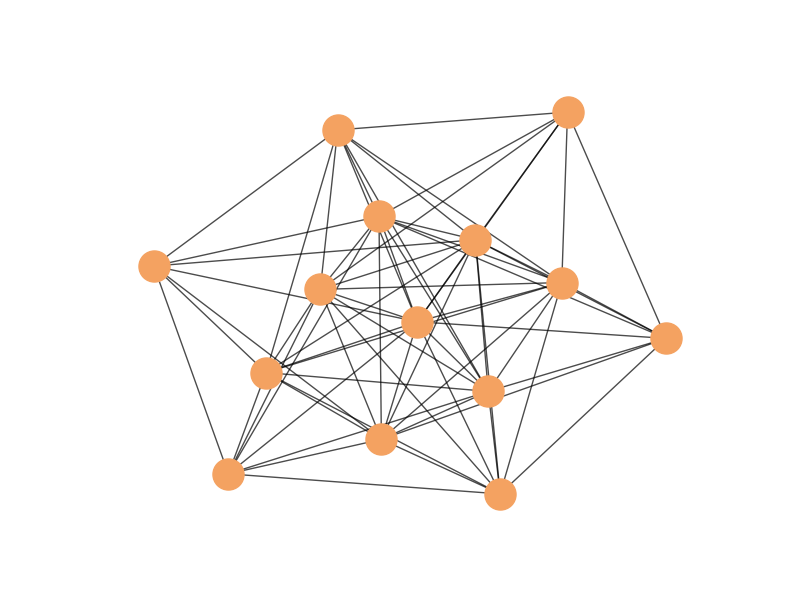}};
        \node (img15) at (7.4, -9) {\includegraphics[width=0.24\textwidth,height=0.18\textwidth]{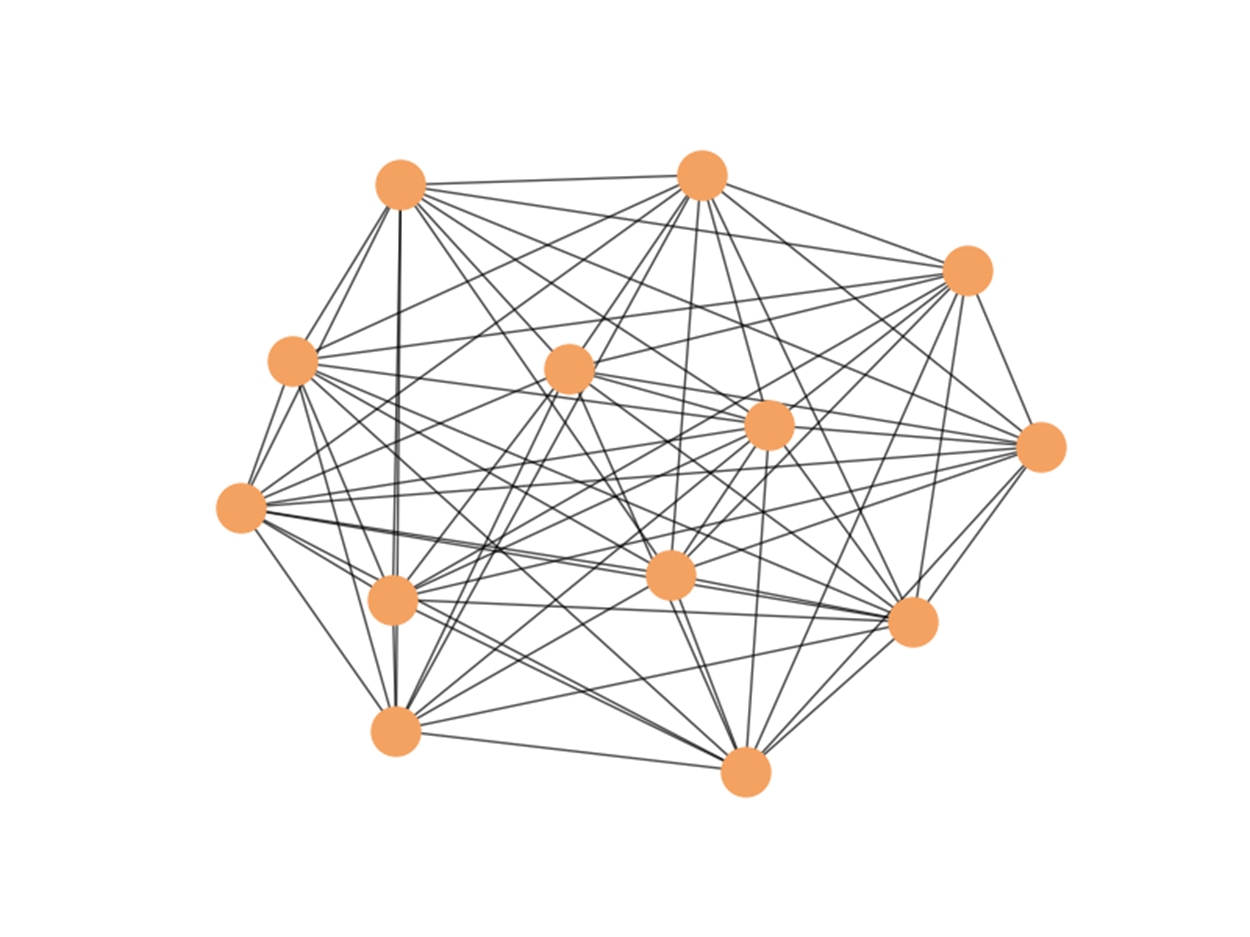}};
        \node (img16) at (11.6, -9) {\includegraphics[width=0.24\textwidth,height=0.18\textwidth]{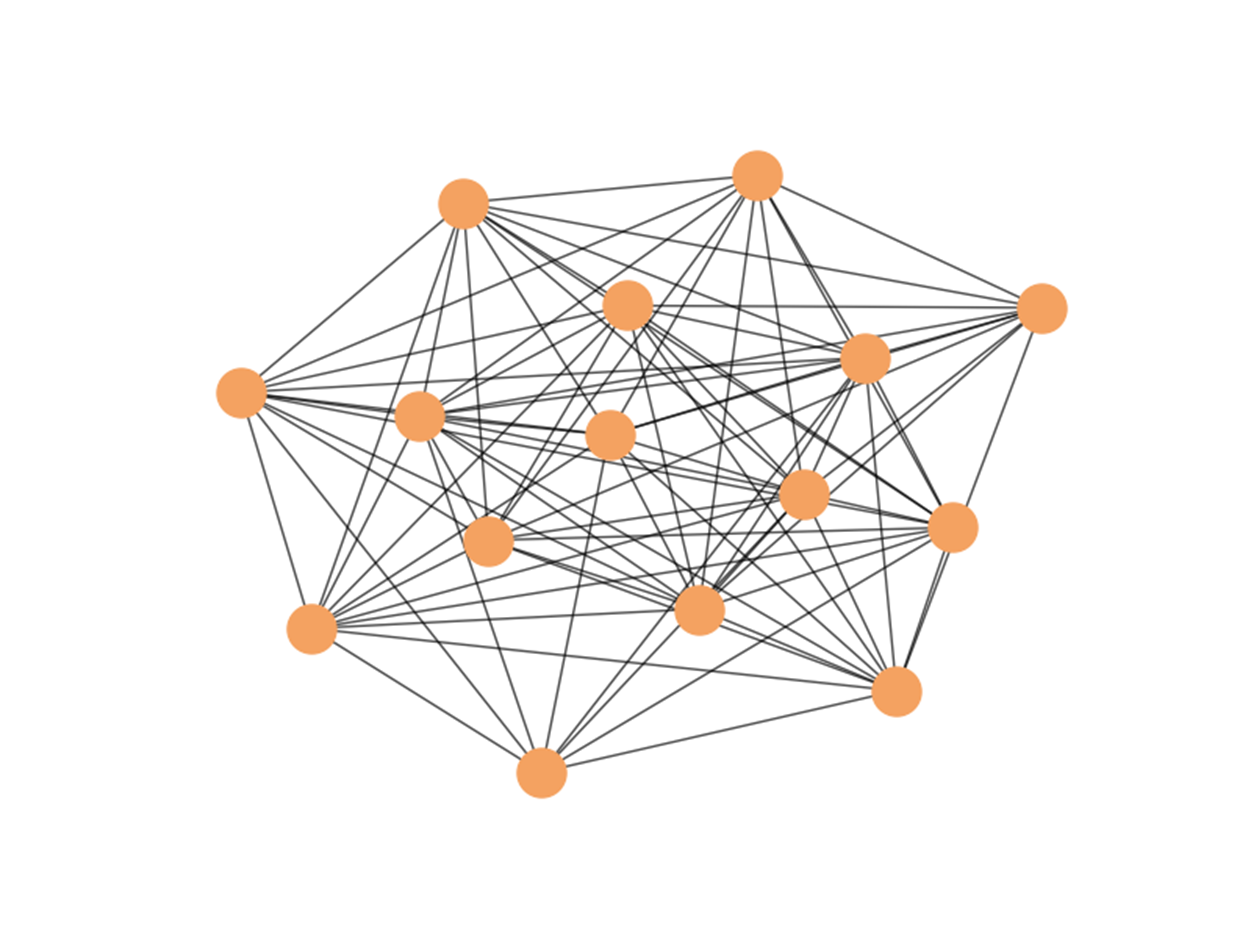}};
        \node[anchor=west, font=\small] at (-4.5, 0) {\textbf{GT:}};
        \node[anchor=west, font=\small] at (-4.5, -3) {\textbf{GFKD:}};
        \node[anchor=west, font=\small] at (-4.5, -6) {\textbf{GFAD:}};
        \node[anchor=west, font=\small] at (-4.5, -9) {\textbf{ACGKD:}};
    \end{tikzpicture}
    \caption{Graph visualization on IMDB-B.}
    \vspace{-8pt}
    \label{imdb-vis}
\end{figure*}
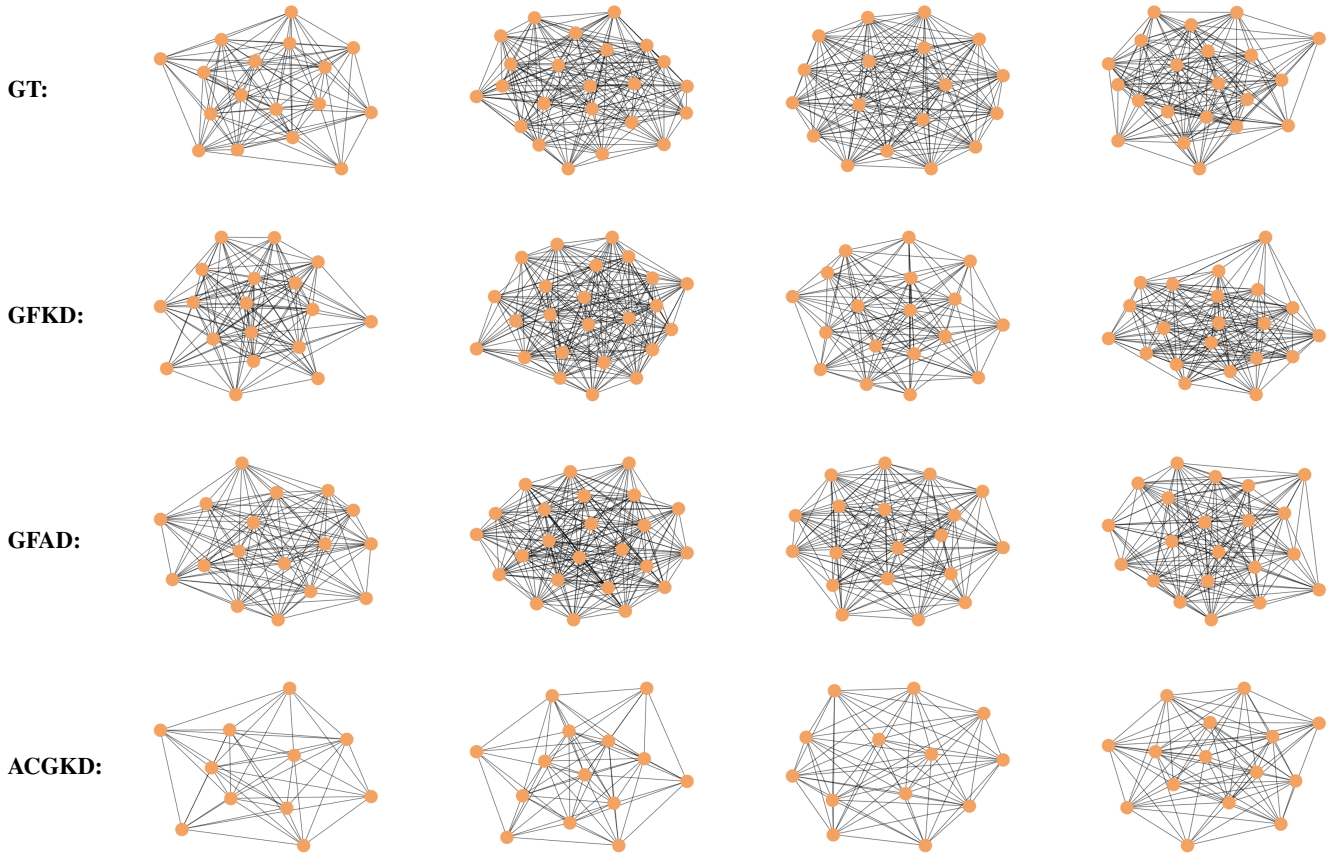

\vspace{0.5em}  
\noindent\textbf{\small Experiments on Social Network Graph Data}
\vspace{0.15em}    

To futher investigate the generalization capability of ACGKD, we conduct experiments on three social network graph datasets. Table \ref{social} presents the comparison results. Note that for these three datasets, the node features are either node degrees or constants, derived from the graph structures. As a result, DG reduces to RG. It can be observed that, except for the COLLAB dataset, ACGKD outperforms the baselines on the other two datasets, demonstrating its strong generalization ability across different types of graph data. The superiority of ACGKD lies in its ability to simplify graph structures while preserving key structural information, reusing the teacher classifier, and introducing adversarial curriculum learning.

\noindent\textbf{\small Comparison of Data Generation Time}
\vspace{0.15em} 

To quantify the impact of node number optimization, we record the data generation time for both GFKD and our ACGKD teacher models. To ensure a fair comparison, we set the total number of iterations for data generation to 1800 and select GCN-5-64 and GIN-5-64 as the teacher models. As shown in Fig. \ref{time-table1} and Fig. \ref{time-table2} (-C represents GCN, and -I represents GIN), across these six datasets, ACGKD reduces time by an average of 58.03\% compared to GFAD and 47.12\% compared to GFKD on GCN architecture, and by 60.25\% compared to GFAD and 49.27\% compared to GFKD on GIN architecture. This demonstrates that our approach effectively reduces training costs by lowering the complexity of graph structure while maintaining the quality of the generated graphs.

\vspace{-3pt}
\subsection{Ablation Studies}
We conduct comprehensive ablation studies on our design. Table \ref{ablation} shows the results of our ablation studies on Curriculum Learning (CL), Dynamic Temperature (DT), and Classifier Reuse (CR).

\noindent\textbf{\small Ablation Study on Curriculum Learning}

We remove both the difficulty regulator, which dynamically adjusts the difficulty of the data generated by the teacher model, and the difficulty control vector used by the student model, while keeping all other components unchanged. Without these two essential elements of curriculum learning, ACGKD's performance significantly decline. This is because curriculum learning allows the student model to be trained progressively from easier to harder tasks, rather than randomly receiving batches of pseudo-graphs, which would otherwise lead to extensive trial-and-error in parameter optimization.

\vspace{0.25em} 
\noindent\textbf{\small Ablation Study on Dynamic Temperature}

To verify the effectiveness of the dynamic temperature, we replace the dynamic temperature with a fixed value $\tau$, and set $\tau$ = 2 based on prior experience from \cite{ctkd}. The experimental results show that, compared to the fixed temperature, the dynamic temperature generally improves the model's performance. This demonstrates the effectiveness of the dynamic temperature in graph-free tasks, as its adversarial mechanism enhances the student model's generalization ability.

\begin{figure*}[htbp]
    \centering
    \subfloat[Teacher]{\includegraphics[width=0.23\linewidth]{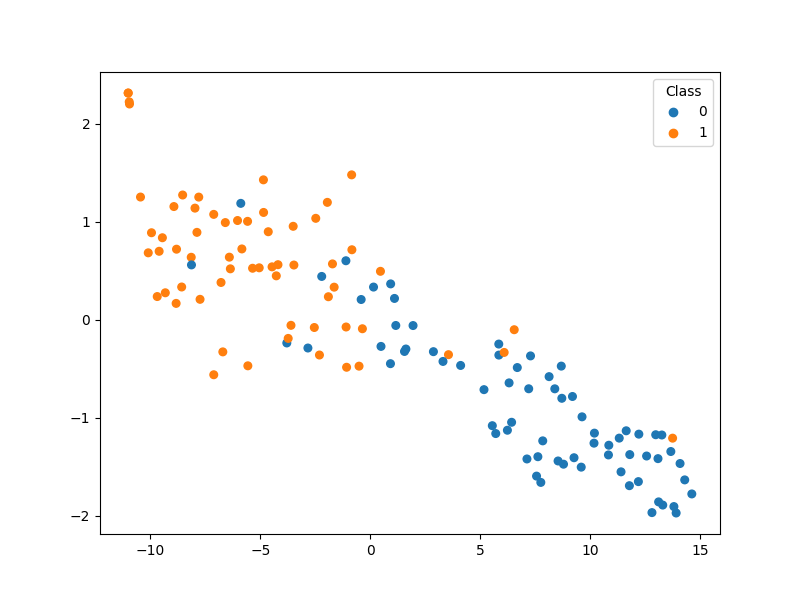}%
    \label{fig:teacher-mutag}}
    \hfil
    \subfloat[GFKD]{\includegraphics[width=0.23\linewidth]{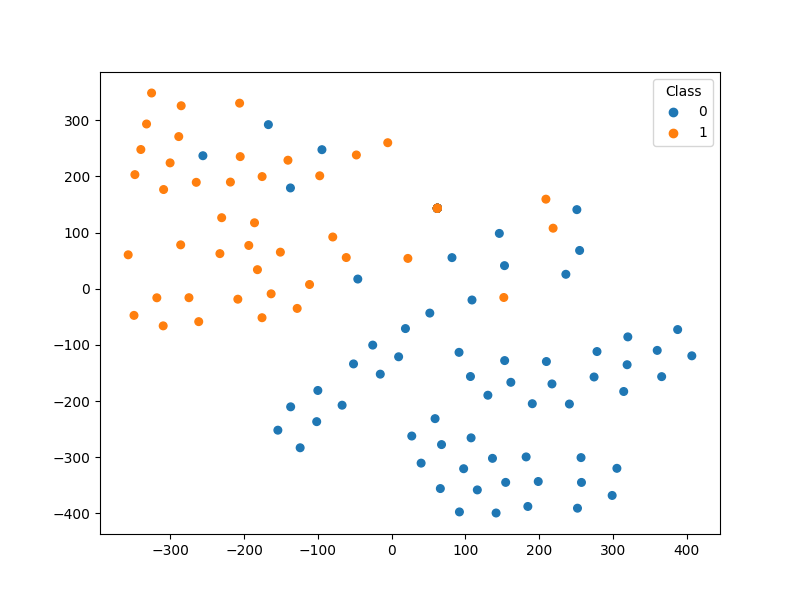}%
    \label{fig:gfkd-mutag}}
    \hfil
    \subfloat[GFAD]{\includegraphics[width=0.23\linewidth]{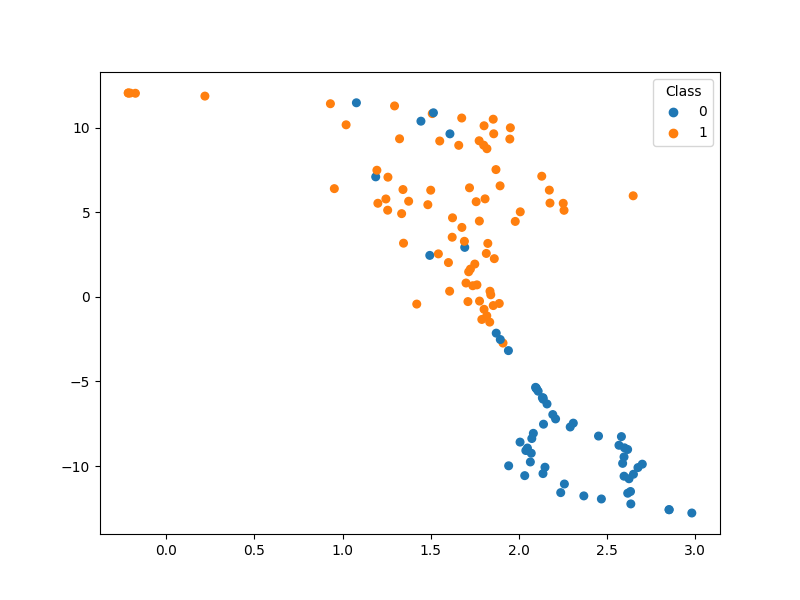}%
    \label{fig:gfad-mutag}}
    \hfil
    \subfloat[ACGKD]{\includegraphics[width=0.23\linewidth]{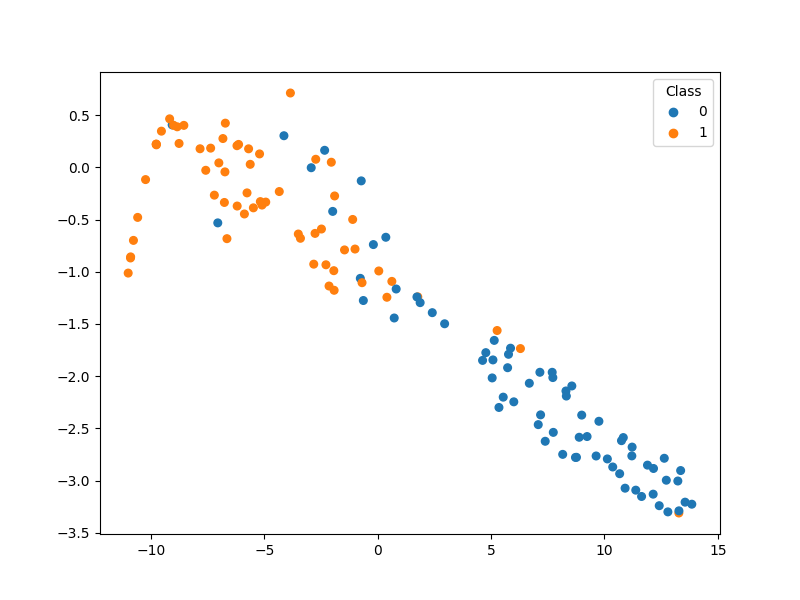}%
    \label{fig:acgkd-mutag}}
    \caption{Feature visualization on MUTAG}
    \label{fig:mutag_visualization}
    \vspace{-10pt}
\end{figure*}

\begin{figure*}[htbp]
    \centering
    \subfloat[Teacher]{\includegraphics[width=0.23\linewidth]{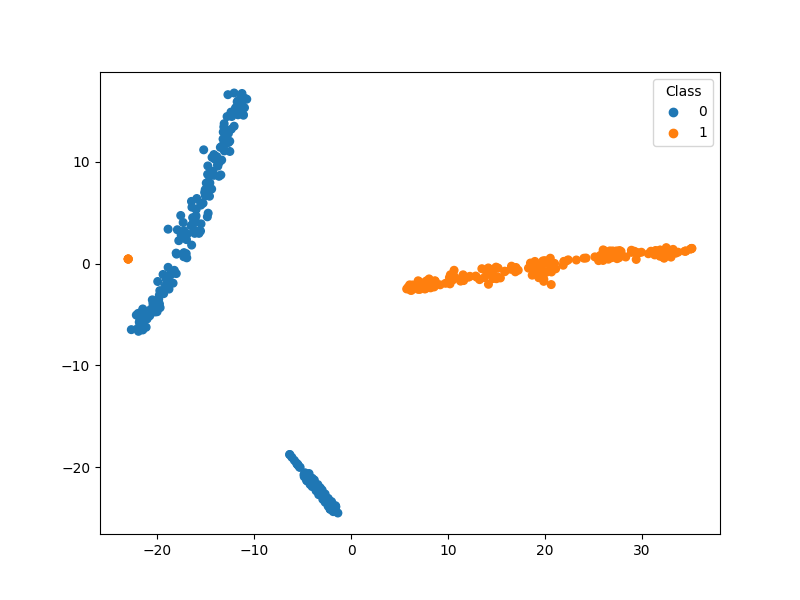}%
    \label{fig:teacher-imdb}}
    \hfil
    \subfloat[GFKD]{\includegraphics[width=0.23\linewidth]{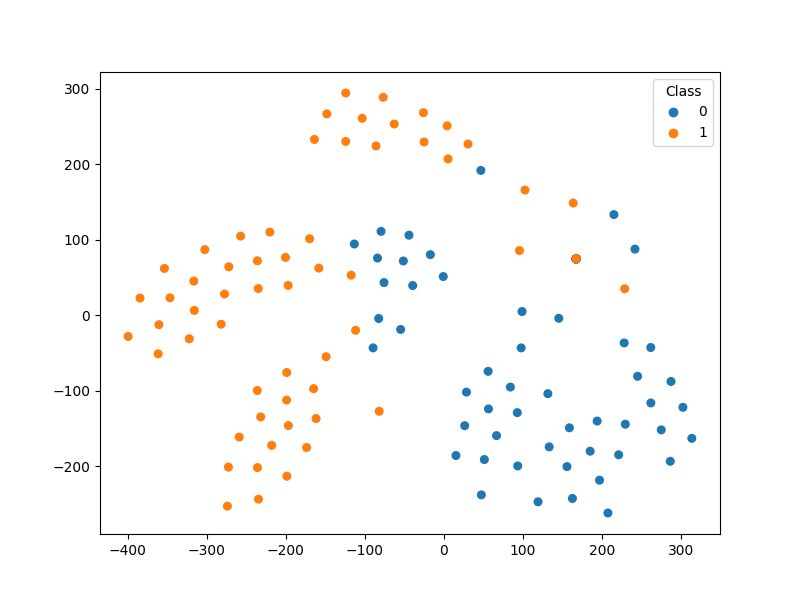}%
    \label{fig:gfkd-imdb}}
    \hfil
    \subfloat[GFAD]{\includegraphics[width=0.23\linewidth]{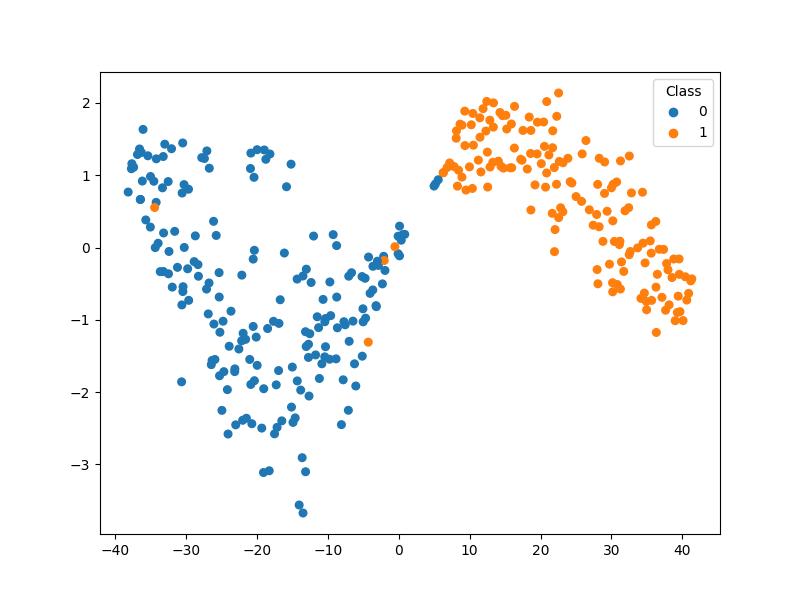}%
    \label{fig:gfad-imdb}}
    \hfil
    \subfloat[ACGKD]{\includegraphics[width=0.23\linewidth]{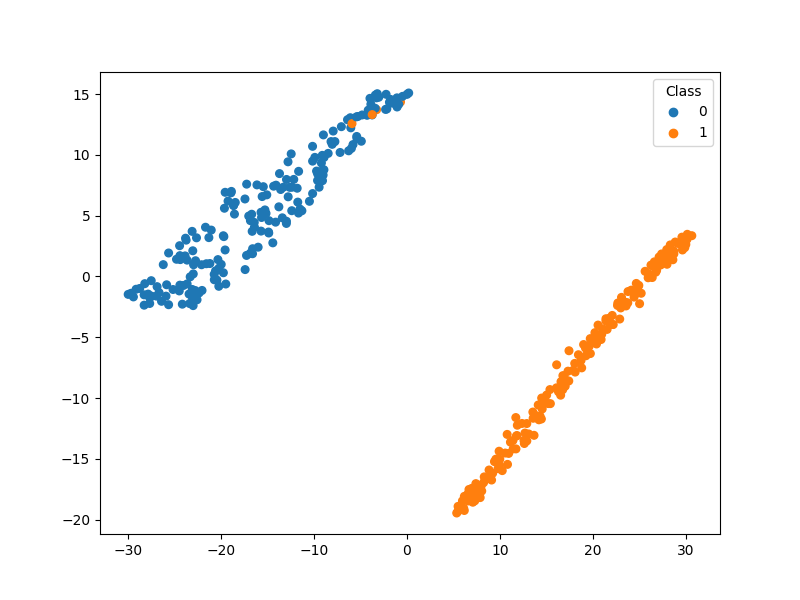}%
    \label{fig:acgkd-imdb}}
    \caption{Feature visualization on IMDB-B}
    \label{fig:imdb_visualization}
\end{figure*}

\vspace{0.25em} 
\noindent\textbf{\small Ablation Study on Classifier Reuse}

We remove the projector used for elevating the student's output to a higher dimension and train a dedicated classifier for the student. It can be observed that, compared to the full ACGKD model, the performance slightly decreases on most datasets when the teacher's classifier is no longer reused. This demonstrates the effectiveness of reusing the classifier. However, it is worth mentioning that, since the independent student model is trained from scratch, this experiment maximally reflects the impact of reducing spatial complexity on graph-free knowledge distillation. Notably, the independently trained student model still achieves efficient and stable performance, even excelling in the three-class COLLAB dataset.

\vspace{-5pt}
\subsection{Graph Data Visualization}
We present some pseudo-samples generated by the teacher over a complete cycle in Figure \ref{curriculum}. It can be observed that, as time progresses, the complexity of the topology generated by the teacher model gradually increases. This demonstrates that the teacher follows an easy-to-hard knowledge transfer order, which is a direct manifestation of curriculum learning.

We also visualize the original datasets, the pseudo-graphs generated by GFKD, GFAD and ACGKD. As shown in Fig. \ref{imdb-vis}, while GFKD and GFAD can generate relatively similar pseudo-graphs, it does not fundamentally simplify the graph structure. In contrast, ACGKD retains essential features while reducing graph spatial complexity. This approach significantly reduces training costs and distillation time.

To further explore whether ACGKD can learn discriminative features from the pseudo-graphs, we utilize t-SNE \cite{r4} to visualize the features learned by different methods. The visualization experiments are conducted on the MUTAG and IMDB-B datasets, with GCN-5-64 as the teacher model and GCN-3-32 as the student model.

As shown in Fig. \ref{fig:mutag_visualization} and Fig. \ref{fig:imdb_visualization}, we visualize the output features of the Teacher, GFKD, GFAD and our ACGKD, with the Teacher serving as a reference. It can be observed that while GFKD shows some degree of feature separation between different classes, the level of feature clustering is not ideal. GFAD is generally able to distinguish between classes, but the degree of clustering is still insufficient. In contrast, the features learned by ACGKD are clearly separated by class, with even better feature clustering than the Teacher. This demonstrates that the pseudo-graphs ACGKD generates have a strong capability to represent graph structures, and our distillation technique has achieved excellent results.

\section{Conclusion}
This paper addresses the limitations of existing graph-free KD methods, including complex structure gradient computation, the possibility of overlooking key edge information, and high spatial complexity of the generated graphs. We propose ACGKD, a novel approach that optimizes the computation of structural gradients and significantly reduces the spatial complexity of graph data, thereby greatly reducing distillation time while adopting a CL-based strategy for the student model. Specifically, we use the Binary Concrete distribution to simplify structural gradient computation while preserving key edge information and introduce a spatial complexity parameter to optimize graph structures. We also enhance the student's output dimensions through GAT and reuse the teacher model's classifier, leading to the student's ability to aggregate graph class information that approaches or even exceeds that of the teacher. To further improve the student's performance, we gradually increase the difficulty of pseudo-samples and use a dynamic vector to control the focus of the student's learning. Additionally, a dynamic temperature is applied to encourage adversarial learning. Extensive experiments on six benchmark datasets demonstrate the superiority of ACGKD in extracting knowledge from GNNs without observable graphs. We hope our work can provide some inspiration for further explorations in graph-free knowledge distillation. Our method still has some limitations, particularly in its performance on three-class datasets (e.g., COLLAB), where there is room for improvement. This will be a focus for future optimization. 

\section{ACKNOWLEDGMENTS}
We thank the anonymous reviewers for their constructive and helpful reviews. This work was supported by National Natural Science Foundation of China Project No.62361166629, No.62176188, No.623B2080, No.623B2086 and No.U21A20427, the Science \& Technology Innovation 2030 Major Program Project No2021ZD0150100, Project No.WU2022A009 from the Center of Synthetic Biology and Integrated Bioengineering of Westlake University, Project No.WU2023C019 from the Westlake University Industries of the Future Research and Key Research and Development Project of Hubei Province (2022BAD175). The numerical calculations in this paper had been supported by the super-computing system in the Supercomputing Center of Wuhan University.


\end{document}